\pdfoutput=1
\documentclass[11pt]{article}

\usepackage[final]{acl}

\usepackage{times}
\usepackage{latexsym}

\usepackage[T1]{fontenc}
\usepackage[utf8]{inputenc}

\usepackage{microtype}

\usepackage{graphicx}
\usepackage{caption}
\usepackage{subcaption}
\usepackage{amsmath}

\usepackage{booktabs}
\usepackage{tabularx}
\usepackage{multirow}
\usepackage{placeins}
\usepackage{rotating}
\usepackage{makecell}
\usepackage{enumitem}
\usepackage{listings} % 문서의 preamble에서 선언
\usepackage{tcolorbox}
\usepackage[table]{xcolor}
\definecolor{lightgray}{gray}{0.9}
\usepackage{todonotes}

\title{What Models Know, How Well They Know It:
Knowledge-Weighted Fine-Tuning for Learning When to Say "I Don't Know"}

\author{
  \textbf{Joosung Lee}$^1$ \quad
  \textbf{Hwiyeol Jo}$^1$ \quad
  \textbf{Donghyeon Ko}$^1$ \quad
  \textbf{Kyubyung Chae}$^2$ \quad \\
  \textbf{Cheonbok Park}$^{1,3}$ \quad  
  \textbf{Jeonghoon Kim}$^{1,3}$\textsuperscript{$\dagger$} \\
  $^1$NAVER CLOUD\quad$^2$Seoul National University\quad$^3$KAIST\\
  \texttt{\{rung.joo,jeonghoon.samuel\}@navercorp.com}
}

\begin{document}
\maketitle
\begin{abstract}
While large language models (LLMs) demonstrate strong capabilities across diverse user queries, they still suffer from hallucinations, often arising from knowledge misalignment between pre-training and fine-tuning.
To address this misalignment, we reliably estimate a fine-grained, instance-level knowledge score via multi-sampled inference.
Using the knowledge score, we scale the learning signal according to the model’s existing knowledge, while encouraging explicit “I don’t know” responses for out-of-scope queries.
Experimental results show that this approach allows the model to explicitly express uncertainty when it lacks knowledge, while maintaining accuracy on questions it can answer.
Furthermore, we propose evaluation metrics for uncertainty, showing that accurate discrimination between known and unknown instances consistently improves performance.
\end{abstract}

\renewcommand{\thefootnote}{\fnsymbol{footnote}}
\footnotetext[2]{Corresponding author.}

\section{Introduction}

LLMs perform effectively on many instruction-based tasks but still struggle with persistent hallucinations, often generating confident yet incorrect content.
Recent studies~\cite{NEURIPS2024_14c018d2, gekhman-etal-2024-fine, kang-etal-2025-unfamiliar} identify a key cause of these hallucinations as knowledge misalignment between the parametric knowledge acquired during pre-training and the supervised fine-tuning datasets. 
When fine-tuning datasets contain knowledge that was rarely or never encountered during pre-training, models are compelled to produce labeled tokens even for facts they inherently do not "know." 
As a result, models tend to generate answers for unanswerable queries instead of explicitly acknowledging the absence of required knowledge, which directly leads to hallucinations.
This suggests that fine-tuning on unknown data makes the model more likely to generate hallucinations.

\begin{table}[t]
\centering
\resizebox{0.8\columnwidth}{!}{%
\begin{tabular}{m{0.5\columnwidth} m{0.24\columnwidth} m{0.24\columnwidth}}
\toprule
\textbf{Question} 
& \textbf{\shortstack{SFT \\ Response}} 
& \textbf{\shortstack{KWT \\ Response}} \\
\midrule
Which genus contains more species: Fallopia or Nicandra?
& {\color{blue}Fallopia}
& {\color{blue}Fallopia} \\
\midrule
Where was the host of Pressure Pad born?
& {\color{red}Glasgow, Scotland}
& {\color{red}Glasgow, Scotland} <IDK> \\
\bottomrule
\end{tabular}
}
\caption{
Comparison of response behaviors with and without instance-level knowledge awareness.
SFT generates confident but incorrect answers, whereas KWT generates <IDK> to express uncertainty.
Blue and red text indicate correct and incorrect answer content, respectively.
}
\label{tab:cherry_pick_append_idk}
\vspace{-3mm}
\end{table}

Recent studies encourage uncertainty expression by training models to distinguish known from unknown data.
\textbf{R-Tuning}~\cite{rtuning} trains the model by categorizing instances into known and unknown based on estimated knowledge, encouraging linguistic expressions of uncertainty for unknown instances.
However, prior works~\citep{gekhman-etal-2024-fine, zhong-etal-2025-kaft} show that the degree to which a model knows a fact substantially influences how the model learns, indicating that knowledge is better represented as a graded quantity rather than a binary one.
\textbf{IDK-Tuning}~\cite{NEURIPS2024_14c018d2} and \textbf{SEAL}~\cite{seal} instead allocate probability mass to a special refusal token to reject predictions that are inconsistent with the model’s knowledge distribution.
However, generating a refusal token during decoding enforces token-level decisions, leading to an inherent trade-off between accuracy and refusal behavior.

\citet{pik} suggests that LLMs exhibit latent self-knowledge through internal signals such as $\mathrm{P(True)}$ and $\mathrm{P(IK)}$.
However, determining whether a model truly knows a fact based on a single response to a question can be unreliable.
Therefore, our goal is to estimate the model’s knowledge in a fine-grained manner and to train it to explicitly express “I don’t know” when relevant knowledge is absent.
Before introducing our method, we briefly illustrate why distinguishing knowledge during training is important.
As shown in Table~\ref{tab:cherry_pick_append_idk}, SFT generates answers regardless of whether the required knowledge is known, whereas KWT expresses uncertainty by generating <IDK> when the knowledge is absent.
Specifically, we estimate knowledge by performing multi-sampled inference over the training data and computing a knowledge score based on the accuracy of generated responses.
This score is then used as a fine-tuning signal, allowing the training strength to vary according to the model’s existing knowledge.
As a result, the model adjusts its training based on what it already knows and explicitly expresses uncertainty when its knowledge is insufficient.

We evaluate our approach on diverse question-answering datasets spanning general, medical, and scientific knowledge.
Across the in-domain evaluations, our knowledge-weighted fine-tuning achieves accuracy comparable to standard supervised fine-tuning, while improving uncertainty expression on unknown queries.
To further test generalization, we also evaluate out-of-domain datasets that explicitly separate answerable and unanswerable queries.
In out-of-domain evaluations, our method improves the model’s ability to express uncertainty, enabling more reliable handling of questions beyond its parametric knowledge.

Our main contributions are as follows:
\begin{itemize}[nosep, leftmargin=1em]
    \item We show that estimating how well the model knows each instance is important for reducing hallucinations. To this end, we conduct experiments using multi-sampled inference and multiple answer-matching metrics to measure the model’s knowledge more reliably.
    
    \item We demonstrate that how the model leverages this knowledge score during fine-tuning (e.g., through knowledge-dependent weighting strategies) plays a critical role in improving both uncertainty expression and overall performance.
    
    \item We introduce evaluation metrics to systematically measure knowledge awareness and uncertainty expression.
\end{itemize}

\section{Related Works}

\subsection{Mitigating Hallucination}
Approaches to reducing hallucinations in LLMs can be grouped into three categories. 
First, some methods use additional supervision. ~\citet{tian2024finetuning} assigns hallucination scores and constructs preference pairs for DPO training, while ~\citet{li-etal-2025-know} builds an uncertainty-aware MRC dataset that labels answerability to enable more reliable instruction tuning.
Second, decoding-based methods reduce hallucinations by comparing multiple logit sources. ~\citet{li-etal-2023-contrastive} contrasts expert and amateur models, ~\citet{chuang2024dola} compares logits from intermediate layers, and ~\citet{shi-etal-2024-trusting} contrasts logits with and without contextual evidence to select more factual next tokens.
Third, test-time scaling approaches regenerate or select outputs at inference. Methods such as ~\citet{madaan2023selfrefine} and ~\citet{mndler2024selfcontradictory} iteratively refine initial answers using self-feedback, while other approaches~\citep{chen2023universalselfconsistency, wang-etal-2024-integrate, wang-etal-2024-improving} generate multiple candidate responses and choose the most reliable one using self-consistency-style selection.

\subsection{Knowledge-Aware Tuning}
\textbf{FT-TOP}~\cite{popular} reports that filtering the training data to retain only samples that the pre-trained model is estimated to already know leads to improved model performance.
\citet{ren-etal-2024-learning} find that instruction fine-tuning is most effective when training data aligns with a model’s existing internal knowledge, and that injecting inconsistent knowledge can degrade performance.
\textbf{KaFT}~\citep{zhong-etal-2025-kaft} detects conflicts between a model’s parametric knowledge and supervision data and assigns discrete weights to improve domain-specific multiple-choice QA. Since KaFT operates in a multiple-choice setting, correctness is directly determined by option selection, without addressing semantic matching challenges in free-form generation.

We focus on reducing hallucinations by addressing knowledge misalignment between pre-training and fine-tuning.
We study open-ended question answering by estimating instance-level knowledge as a fine-grained value to guide fine-tuning.
By using a special token to express uncertainty on unknown instances, our method supports uncertainty-aware responses while remaining compatible with standard answer generation.

\section{Method}

We conduct experiments on question–answering datasets.
Each fine-tuning sample is represented as $(q, r^\ast, k)$, where $q$ is the input question, $r^\ast$ is the gold response, and $k$ is an optional supporting knowledge paragraph when such information is available.
We denote the pre-trained model as $M_{\text{base}}$.
Given a question $q$, $M_{\text{base}}$ generates a prediction $\hat{r}$, which is compared against the gold response $r^\ast$ to estimate the model’s knowledge of the instance.
When available, the knowledge paragraph $k$ is used for correctness evaluation under the LLM-as-a-judge framework~\citep{zheng2023judging}.

To obtain a reliable estimate of the model’s knowledge, we infer multiple responses rather than relying on a single $\hat{r}$, and assess their consistency with the gold response via multi-sampled inference.
The resulting fine-grained knowledge score is used to modulate the fine-tuning signal, so that each instance is trained in proportion to the model’s prior knowledge.
For instances with insufficient knowledge, we additionally encourage the model to generate a special <IDK> token.
As a result, the model learns to answer confidently when sufficient knowledge is present, while explicitly refusing to answer by producing <IDK> when the required knowledge is absent.

\begin{table}[t]
\centering
\resizebox{0.9\columnwidth}{!}{%
\begin{tabular}{lcccc}
\toprule
\textbf{Metric} & \textbf{KP-Acc} & \textbf{KP-F1} & \textbf{KNP-Acc} & \textbf{KNP-F1} \\
\midrule
EM     & 63.2 & 57.2 & 71.2 & 55.8 \\
Rouge  & 82.4 & 85.8 & 83.8 & 81.6 \\
LLM-as-a-judge    & 93.4 & 94.8 & 93.8 & 93.4 \\
\bottomrule
\end{tabular}%
}
\caption{Agreement of automatic metrics (EM, ROUGE, LLM) with human evaluation under knowledge-provided (KP) and knowledge-not-provided (KNP) settings.}
\label{tab:knowledge_comparison}
\vspace{-3mm}
\end{table}

\begin{table*}[t]
\centering
\resizebox{0.85\textwidth}{!}{%
\begin{tabular}{c *{3}{ccc}}
\toprule
 & \multicolumn{3}{c}{\textbf{EM}} & \multicolumn{3}{c}{\textbf{Rouge}} & \multicolumn{3}{c}{\textbf{LLM}} \\
\cmidrule(lr){2-4} \cmidrule(lr){5-7} \cmidrule(lr){8-10}
\textbf{KnowledgeScore (KS)} 
& \textbf{HaluEval} & \textbf{MedQA} & \textbf{SciQ}
& \textbf{HaluEval} & \textbf{MedQA} & \textbf{SciQ}
& \textbf{HaluEval} & \textbf{MedQA} & \textbf{SciQ} \\
\midrule
0 & 66.75 & 87.98 & 40.00 & 49.83 & 80.28 & 25.84 & 47.74 & 67.34 & 14.38 \\
0.2 &  9.34 &  6.60 & 12.66 & 12.59 & 10.26 & 11.97 & 13.13 & 15.74 &  9.07 \\
0.4 &  7.24 &  2.63 &  9.58 &  9.41 &  4.45 &  9.80 &  8.80 &  7.69 &  9.21 \\
0.6 &  5.85 &  1.56 &  9.21 &  8.48 &  2.63 & 10.75 &  8.55 &  4.75 & 10.79 \\
0.8 &  5.68 &  0.87 & 10.52 &  8.76 &  1.58 & 13.21 &  9.34 &  2.77 & 15.93 \\
1.0 &  5.15 &  0.34 & 18.02 & 10.94 &  0.80 & 28.44 & 12.45 &  1.71 & 40.62 \\
\bottomrule
\end{tabular}
}
\caption{Distribution (\%) of questions across knowledge score bins, computed from multi-sampled model responses generated by Llama 3.2-3B and evaluated using EM, ROUGE, and LLM-based metrics.}
\label{tab:correct_dist}
\vspace{-3mm}
\end{table*}

\subsection{Knowledge Estimation}
\label{sec:knowledge_estimation}
Even if $M_{\text{base}}$ has parametric knowledge of a fact, it may not reliably express that knowledge during generation. 
Although “not generated” and “not known” are conceptually distinct, following prior works~\cite{gekhman-etal-2024-fine, yang2024alignment, rtuning}, we operationalize knowledge as the ability to produce a correct answer under typical inference conditions, and treat failures to do so as effectively unknown for fine-tuning.
We compute the model's knowledge not as a binary known-or-unknown classification, but as a fine-grained score derived through the following steps.

\paragraph{Multi-sampled few-shot inference for robust knowledge estimation.}
Because pre-trained models may not reliably follow instructions, we probe knowledge using 3-shot prompting. For each question, we sample three demonstration QA pairs to form a 3-shot prompt and then sample one response. We repeat this process with independently resampled demonstrations $S=5$ times, yielding responses under diverse prompting conditions. Inspired by self-consistency~\citep{wang2023selfconsistency}, this produces a more robust estimate of the model’s parametric knowledge.

\paragraph{Measuring correctness by comparing sampled responses with gold answers.}
Each sampled response is evaluated against the gold answer to determine whether it is correct.
We attempt to compare the two responses using the following three methods:
Exact Match (EM) to detect strictly correct predictions but may miss semantically correct answers with different surface forms. Following~\citet{kuhn2023semantic}, we deem a response correct if its ROUGE score exceeds a certain threshold, allowing lexically varied but semantically aligned answers to be counted as correct. 
Lastly, we use an LLM-as-a-judge, which compares the response and the gold answer to determine semantic correctness.
The prompts for evaluating the LLM are described in Appendix~\ref{sec:llm-as-a-judeg-prompt}.

Consequently, we compute a fine-grained knowledge score by averaging the correctness of the sampled responses. Given generated responses ($\{\hat{r}_i\}_{i=1}^S$) and a matching function ($f(\cdot)$) that determines whether a response matches the gold response, the knowledge score is defined as:

{\small
\begin{equation}
\text{KnowledgeScore}(q) = \frac{1}{S} \sum_{i=1}^{S} f\big(\hat{r}_i, r^\ast\big)
\end{equation}
}

To assess the reliability of these automatic correctness metrics, we obtain human judgments from a single annotator for 100 questions (500 responses) drawn from both a knowledge-provided dataset (HaluEval) and a knowledge-not-provided dataset (MedQA).
For the knowledge-provided dataset, the relevant knowledge paragraph is included in the prompt when evaluating responses with the LLM-as-a-judge method.
As shown in Table~\ref{tab:knowledge_comparison}, LLM-as-a-judge exhibits the highest agreement with human judgments, followed by ROUGE and EM, across both datasets.
The ROUGE threshold is selected based on the configuration that yields the highest agreement with human judgments.
These results indicate that LLM-as-a-judge is the most suitable matching function.
Unless otherwise specified, we therefore compute knowledge scores using the LLM-as-a-judge metric and evaluate all reported performance metrics using the same criterion.

\subsection{Knowledge-Weighted Fine-Tuning}
\label{sec:knowledge_strategy}
We assign sample-specific training weights using the instance-level knowledge scores, whose distribution is reported in Table~\ref{tab:correct_dist}. 
Specifically, $w_i$ denotes the training weight assigned to the $i$-th sample, which modulates the contribution of each instance to the overall loss during fine-tuning:

{\small
\begin{equation}
\mathcal{L}
= - \frac{1}{NM} \sum_{i=1}^{N} w_i \sum_{t=1}^{M} \log p(x_t^{(i)} \mid x_{<t}^{(i)}).
\end{equation}
}
where $x^{(i)} = (x^{(i)}_1, \ldots, x^{(i)}_M)$ is the target token sequence of the $i$-th training sample, 
$N$ denotes the total number of training samples, and $M$ denotes the number of target tokens in each sample.

To understand how the knowledge score relates to effective fine-tuning, we explore three weighting strategies:
(1) \textbf{Familiarity (F)} weighting ($w^{f}$), which increases the contribution of samples the model already knows well; 
(2) \textbf{Reverse-Familiarity (RF)} weighting ($w^{rf}$), which down-weights well-known samples; and 
(3) \textbf{Uniform (U)} weighting ($w^{u}$), which keeps all weights identical.
Specifically, we use:

{\small
\begin{equation}
w_i^{f} = \frac{S \times \mathrm{KS}_i + 1}{S+1},
\end{equation}
\begin{equation}
w_i^{rf} = 1 - \frac{S \times \mathrm{KS}_i}{S+1},
\end{equation}
\begin{equation}
w_i^{u} = 1.0, 
\end{equation}
}

where $KS_{i}$ represents the knowledge score of the $i$-th sample and $S$ denotes the number of sampled responses.
We adopt the F strategy as our default weighting strategy. The effect of different weighting strategies is discussed in Section~\ref{sec:weighting_strategy}.

For samples with a knowledge score of zero, which indicates that the model has no knowledge of the underlying fact, we append a special <IDK> token (representing “I don’t know”) to the end of the target response to encourage explicit uncertainty.
In Appendix~\ref{app:var_response}, this approach is shown to be simple yet effective, with appending <IDK> yielding better performance than replacing the entire response with <IDK>.
Unlike full replacement, our method preserves performance on answerable questions while still promoting appropriate uncertainty expression.
At inference time, users can choose to allow the <IDK> token to enable explicit uncertainty expression, or suppress it to use the model as a standard answer-generating LLM.

\section{Experiments}

\subsection{Implementation Details}
We conduct all experiments using the pretrained Llama 3.2-3B~\cite{grattafiori2024llama3herdmodels}. 
For KWT (ROUGE), we use ROUGE-L F1 as the matching metric. The correctness thresholds are set to 0.35, 0.6, and 0.6 for HaluEval, MedQA, and SciQ, respectively.
When estimating knowledge scores via multi-sampled inference, we use a temperature of 0.7 to encourage diverse generations.
The LLM-as-a-judge model used in our experiments is the instructed Gemma-3 12B~\cite{gemmateam2025gemma3technicalreport}.
All models are trained under the same experimental settings for 3 epochs.
To demonstrate that our findings generalize across models, we additionally conduct experiments on Qwen 2.5-3B~\cite{qwen2025qwen25technicalreport} in Appendix~\ref{sec:qwen2.5}.

\subsection{Datasets}
We empirically evaluate our method across diverse open-domain QA datasets, including HaluEval~\cite{halueval} for capturing general hallucination scenarios, MedQA~\cite{medqa} for medical knowledge assessment, and SciQ~\cite{sciq} for scientific commonsense evaluation. 
For HaluEval, we use the provided 10,000 samples and split them into training and test sets with an 8:2 ratio.
Additionally, we evaluated on explicitly answerable/unanswerable datasets such as NEC~\cite{necrefunq}, RefuNQ~\cite{necrefunq}, and SelfAware~\cite{selfaware}, verifying the model's generalized capability to distinguish and appropriately express uncertainty across out-of-domain.

\subsection{Evaluation Metrics}
Evaluating a model’s ability to appropriately respond with uncertainty is non-trivial. 
Standard accuracy only measures answer correctness and does not reward correct abstention.
\citet{kalai2025languagemodelshallucinate} show that standard accuracy-based evaluation inherently penalizes abstention and reward confident guessing, which reinforces hallucination-like behavior rather than discouraging it.
This motivates the use of evaluation metrics that explicitly account for appropriate uncertainty expression, and we therefore adopt the following metrics to capture different aspects of uncertainty-aware behavior.
All evaluation metrics described below are computed based on the four outcome categories (A, B, C, D) defined in Table~\ref{tab:metric_num_table}.

\paragraph{Normalized Area Under the Product Curve (nAUPC).}

Uncertainty-Aware Accuracy (UA-Acc) measures a model’s ability to appropriately abstain on incorrect answers by assigning partial credit:

{\small
\begin{equation}
\mathrm{UA\text{-}Acc}(\alpha)
= \frac{\alpha B + C}{A + B + C + D}
\end{equation}
}
where $\alpha$ controls the reward for abstention on incorrect answers.

Since UA-Acc does not penalize abstention on correct answers, we define the
complementary Certainty-Aware Accuracy (CA-Acc) as:
{\small
\begin{equation}
\mathrm{CA\text{-}Acc}(\alpha)
= \frac{C - \alpha A}{A + B + C + D}
\end{equation}
}
where $\alpha$ penalizes unnecessary abstention.

Varying $\alpha$ induces a continuum of operating points that reflect different balances between uncertainty expression and certainty.
To summarize this trade-off with a single scalar, we define the nAUPC as:

{\small
\begin{equation}
\begin{aligned}
\mathrm{nAUPC}
&=
\frac{1}{\alpha_{\max}}
\int_{0}^{\alpha_{\max}}
\mathrm{UA\text{-}Acc}(\alpha)\,
\mathrm{CA\text{-}Acc}(\alpha)\,
d\alpha
\\
&\approx
\frac{1}{N\,\alpha_{\max}}
\sum_{k=1}^{K}
\Big(
\mathrm{UA\text{-}Acc}(\alpha_{k-1})
\mathrm{CA\text{-}Acc}(\alpha_{k-1})
\\
&\qquad
+
\mathrm{UA\text{-}Acc}(\alpha_k)
\mathrm{CA\text{-}Acc}(\alpha_k)
\Big)
(\alpha_k - \alpha_{k-1}) 
\end{aligned}
\end{equation}
}

where $N = 200$ is a normalization constant accounting for trapezoidal integration and percentage scaling, and $\{\alpha_k\}_{k=0}^{K}$ are uniformly spaced values in $[0,\alpha_{\max}]$.
This definition assigns a high score only when both uncertainty-aware abstention
(UA-Acc) and certainty preservation (CA-Acc) are simultaneously high across the
range of $\alpha$.
We set $\alpha_{\max}=1.0$ in all experiments.

\paragraph{Abstain False Positive Rate (A-FPR).}
The proportion of cases in which the model includes <IDK> in its response even though the answer is correct.
Such cases indicate unnecessary uncertainty expression that may confuse users.

{\small
\begin{equation}
\mathrm{A\text{-}FPR} = \frac{A}{A + C}
\end{equation}
}

\paragraph{IDK Precision.}
IDK Precision quantifies how frequently an <IDK> prediction is justified, measured as the fraction of <IDK> responses that correspond to incorrect answers.

{\small
\begin{equation}
\mathrm{IDK\text{-}Precision} = \frac{B}{A + B}
\end{equation}
}

\begin{table}[t]
\centering
\resizebox{0.35\textwidth}{!}{
\begin{tabular}{c|cc}
\toprule
 & \textbf{Correct} & \textbf{Incorrect} \\
\midrule
\textbf{Response w/ IDK} & A & B \\
\textbf{Response w/o IDK} & C & D \\
\bottomrule
\end{tabular}
}
\caption{Confusion matrix of responses based on correctness and IDK usage.}
\label{tab:metric_num_table}
\vspace{-3mm}
\end{table}

\subsection{Compared Methods}

\textbf{SFT} is standard fine-tuning where $M_{\text{base}}$ is trained without distinguishing whether each sample is known to the model.
In \textbf{FT-TOP}, $M_{\text{base}}$ is fine-tuned only on samples it is estimated to already know.
As a result, neither method is capable of producing IDK responses.
\textbf{R-Tuning} is a method that trains $M_{\text{base}}$ by appending an uncertainty expression to the response on samples it does not know.
For a fair comparison with our method, this uncertainty expression is replaced with the <IDK> token during our experiments.
\textbf{SEAL} trains $M_{\text{base}}$ by comparing the predicted token distribution with the gold token at each step.
When the top-1 prediction does not match the gold token, it reallocates a portion of the one-hot target probability to a special <IDK> token, enabling the model to express uncertainty when its knowledge is unreliable.

\subsection{Results on In-Domain QA}
Table~\ref{tab:main_table} presents the results of our method and baselines.
KWT performance consistently reflects the quality of the underlying knowledge estimation.
KWT (LLM) achieves the strongest uncertainty-aware performance, with higher nAUPC, lower A-FPR, and higher IDK Precision than KWT (ROUGE) and KWT (EM),
demonstrating that more reliable instance-level knowledge estimation leads to better-calibrated uncertainty-aware behavior.
Appending an <IDK> token to samples with zero knowledge score enables precise uncertainty expression without sacrificing accuracy on answerable queries.

\begin{table}[t]
\centering
\resizebox{0.8\columnwidth}{!}{
\begin{tabular}{lccc}
\toprule
\textbf{Model} & \textbf{HaluEval} & \textbf{MedQA} & \textbf{SciQ} \\
\midrule
SFT & \textbf{31.0} & \textbf{20.4} & 62.7 \\
FT-TOP & 29.5 & 17.2 & 63.9 \\
R-Tuning & 30.3 & 19.7 & \textbf{64.6} \\
KWT (EM) & 30.7 & 16.9 & 62.1 \\
KWT (Rouge) & 30.3 & 16.9 & 63.8 \\
\rowcolor{lightgray}
KWT (LLM) & 29.8 & 18.5 & 62.6 \\
\bottomrule
\end{tabular}
}
\caption{Accuracy comparison across datasets, where accuracy is defined as {\small $(A + C) / (A + B + C + D)$}.}
\label{tab:standard_acc}
\vspace{-3mm}
\end{table}

\begin{table}[t]
\centering
\resizebox{1.0\columnwidth}{!}{
\begin{tabular}{l l c c c}
\toprule
\textbf{Dataset} & \textbf{Model} 
& \textbf{nAUPC $\uparrow$}
& \textbf{A-FPR $\downarrow$} 
& \textbf{\makecell{\textbf{IDK} \\ \textbf{Precision}} $\uparrow$} 
\\
\midrule

% ===================== HaluEval =====================
\multirow{4}{*}{HaluEval}
& R-Tuning     & 2.7 & 50.2  & 80.4  \\
& KWT (EM)     & 8.8 & 25.4  & 86.7  \\
& KWT (Rouge)  & 10.2 & 14.5  & 88.9  \\
\rowcolor{lightgray}
& KWT (LLM)    & \textbf{11.3} & \textbf{11.41} & \textbf{92.1} \\
\midrule

% ===================== MedQA =====================
\multirow{4}{*}{MedQA}
& R-Tuning     & -2.0 & 75.3  & 82.9  \\
& KWT (EM)     & 1.3 & 49.3  & 87.6  \\
& KWT (Rouge)  & 2.1 & 37.7  & 88.5  \\
\rowcolor{lightgray}
& KWT (LLM)    & \textbf{4.3} & \textbf{21.6} & \textbf{91.2} \\
\midrule

% ===================== SciQ =====================
\multirow{4}{*}{SciQ}
& R-Tuning     & 9.9 & 44.4  & 47.6  \\
& KWT (EM)     & 26.4 & 19.0  & 62.1  \\
& KWT (Rouge)  & 34.0 & 11.0  & 66.8  \\
\rowcolor{lightgray}
& KWT (LLM)    & \textbf{38.1} & \textbf{3.7} & \textbf{78.3} \\
\bottomrule
\end{tabular}
}
\caption{
Comparison of uncertainty-aware behavior on the in-domain datasets.
\textbf{Bold} indicates the best performance in each column.
}
\label{tab:main_table}
\vspace{-3mm}
\end{table}

\paragraph{Acc.}
Table~\ref{tab:standard_acc} shows that KWT incurs only a slight decrease in standard accuracy compared to SFT.
This result indicates that uncertainty-aware training introduces only a modest trade-off in answer correctness.
A more detailed analysis of the impact of <IDK> supervision on standard accuracy is provided in Appendix~\ref{app:idk_supervision}.
Under the uncertainty-aware evaluation in Table~\ref{tab:main_table}, KWT (LLM) attains the highest nAUPC with lower A-FPR and higher IDK Precision than other methods,
suggesting that instance-level knowledge weighting improves uncertainty expression without degrading standard accuracy compared to R-Tuning.

\paragraph{IDK expression.} 
KWT consistently achieves lower A-FPR and higher IDK Precision than R-Tuning, indicating reduced unnecessary abstention while producing <IDK> responses when appropriate.
This improvement stems from knowledge-weighted fine-tuning, which better distinguishes answerable from unanswerable queries.

\subsection{Results on Out-of-Domain QA}
We further evaluate our method on out-of-domain datasets. Since all models are fine-tuned on the same in-domain data with an identical base model, differences in performance primarily reflect the effect of the fine-tuning strategy rather than domain-specific knowledge advantages.

% As shown in Table~\ref{tab:ood_answerable}, KWT exhibits trends on out-of-domain datasets that are consistent with those observed in the in-domain setting. 
As shown in Table~\ref{tab:ood_answerable}, KWT exhibits similar trends on out-of-domain datasets.
In particular, KWT maintains comparable accuracy while achieving improved uncertainty-aware behavior, indicating that knowledge-weighted fine-tuning generalizes effectively beyond the training distribution. Detailed analysis on explicitly unanswerable queries and the corresponding IDK expression is deferred to Appendix~\ref{app:idk_ratio}.

\begin{table}[t]
\centering
\resizebox{0.46\textwidth}{!}{
\begin{tabular}{l l c c c}
\toprule
\textbf{Dataset} & \textbf{Model} 
& \textbf{nAUPC $\uparrow$}
& \textbf{A-FPR $\downarrow$} 
& \textbf{\makecell{\textbf{IDK} \\ \textbf{Precision}} $\uparrow$}
\\
\midrule

% ===================== RefuNQ =====================
\multirow{4}{*}{RefuNQ}
& R-Tuning     & -1.9 & 73.3 & 81.6 \\
& KWT (EM)     & 2.2 & 45.8 & 86.0 \\
& KWT (Rouge)  & 5.4 & 22.1 & 90.7 \\
\rowcolor{lightgray}
& KWT (LLM)    & \textbf{5.7} & \textbf{19.7} & \textbf{91.0} \\
\midrule

% ===================== SelfAware =====================
\multirow{4}{*}{SelfAware}
& R-Tuning     & -2.0 & 70.7 & 74.7 \\
& KWT (EM)     & 4.6 & 42.3 & 80.5 \\
& KWT (Rouge)  & 8.7 & 19.6 & 85.0 \\
\rowcolor{lightgray}
& KWT (LLM)    & \textbf{9.9} & \textbf{17.1} & \textbf{87.5} \\
\bottomrule
\end{tabular}
}
\caption{
Comparison of uncertainty-aware behavior on the out-of-domain datasets: RefuNQ and SelfAware.
}
\label{tab:ood_answerable}
\vspace{-3mm}
\end{table}

\subsection{Comparison to Token-Level Abstention}
Table~\ref{tab:seal_comparison} compares our method with SEAL, a token-level uncertainty expression approach.
Since SEAL treats any response containing <IDK> as incorrect, it enforces $A=0$, trivially yielding an A-FPR of 0 and an IDK Precision of 100, which makes these metrics uninformative for comparison.

This behavior also limits the interpretability of nAUPC, as nAUPC largely mirrors standard accuracy trends in SEAL rather than reflecting calibrated uncertainty-aware trade-offs.
Under standard accuracy evaluation, SEAL consistently underperforms KWT across in-domain datasets and exhibits substantial performance degradation on out-of-domain benchmarks, indicating limited generalization beyond the training distribution.

\subsection{Effect of Different Weighting Strategies}
\label{sec:weighting_strategy}
We examine not only how the knowledge score should be estimated, but also how it should be utilized during fine-tuning.
We conduct experiments using the alternative weighting strategies introduced in Section~\ref{sec:knowledge_strategy}.

Table~\ref{tab:weighting_diff} reports the results of the reverse-familiarity (RF) and uniform (U) weighting strategies.
While emphasizing low-knowledge samples can provide limited benefits in certain settings, RF and U weighting consistently reduce nAUPC.
A similar trend is observed in Table~\ref{tab:standard_acc}, where SFT is often comparable to or slightly better than FT-TOP, suggesting that some low-knowledge instances still provide useful training signal.
As shown in Table~\ref{tab:correct_dist}, a large fraction of training samples receive zero knowledge scores, and up-weighting such samples can improve answer accuracy in some cases.

However, RF and U consistently result in higher A-FPR and lower IDK Precision across datasets, indicating excessive uncertainty expression even when the model can answer correctly.
Overall, these results suggest that emphasizing low-knowledge samples yields modest nAUPC gains at the cost of degraded uncertainty expression.
In contrast, the familiarity (F) weighting strategy achieves a better balance between answer accuracy and well-calibrated uncertainty expression.

\begin{table}[t]
\centering
\resizebox{0.8\columnwidth}{!}{
\begin{tabular}{l l c c}
\toprule
\textbf{Domain} & \textbf{Dataset} 
& \textbf{Acc $\uparrow$}
& \textbf{nAUPC $\uparrow$} \\
\midrule

% ===================== In-Domain =====================
\multirow{3}{*}{ID}
& HaluEval 
  & 27.9 {\scriptsize(–1.9)} 
  & 11.3 {\scriptsize(0.0)}  \\
& MedQA    
  & 18.3 {\scriptsize(–0.2)} 
  & 5.8 {\scriptsize(+1.5)} \\
& SciQ     
  & 62.6 {\scriptsize(0.0)} 
  & 40.2 {\scriptsize(+2.1)} \\
\midrule

% ===================== Out-of-Domain =====================
\multirow{2}{*}{OOD}
& RefuNQ     
  & 17.5 {\scriptsize(–4.0)} 
  & 5.4 {\scriptsize(-0.3)} \\
& SelfAware  
  & 27.0 {\scriptsize(–3.9)} 
  & 11.1 {\scriptsize(+1.2)} \\
\bottomrule
\end{tabular}
}
\caption{
Performance of SEAL compared to KWT on in-domain (ID) and out-of-domain (OOD) datasets.
Values in parentheses indicate differences from KWT (LLM).
}
\label{tab:seal_comparison}
\end{table}

\begin{table}[t]
\centering
\resizebox{0.48\textwidth}{!}{
\begin{tabular}{l l c c c}
\toprule
\textbf{Dataset} & \textbf{Model} 
& \textbf{nAUPC $\uparrow$}
& \textbf{A-FPR $\downarrow$} 
& \textbf{\makecell{\textbf{IDK} \\ \textbf{Precision}} $\uparrow$} 
\\
\midrule

% ===================== HaluEval =====================
\multirow{2}{*}{HaluEval}
& KWT-RF
  & 9.2 {\scriptsize(-2.1)} 
  & 25.5 {\scriptsize(+14.1)} 
  & 87.1 {\scriptsize(–5.0)} \\
& KWT-U 
  & 9.9 {\scriptsize(-1.4)} 
  & 22.3 {\scriptsize(+10.9)} 
  & 88.1 {\scriptsize(–4.0)} \\
\midrule

% ===================== MedQA =====================
\multirow{2}{*}{MedQA}
& KWT-RF 
  & 2.1 {\scriptsize(-2.2)} 
  & 45.9 {\scriptsize(+24.3)} 
  & 86.5 {\scriptsize(–4.7)} \\
& KWT-U 
  & 2.6 {\scriptsize(-1.7)} 
  & 41.8 {\scriptsize(+20.2)} 
  & 86.5 {\scriptsize(–4.7)} \\
\midrule

% ===================== SciQ =====================
\multirow{2}{*}{SciQ}
& KWT-RF
  & 33.9 {\scriptsize(-4.2)} 
  & 11.1 {\scriptsize(+7.4)} 
  & 70.3 {\scriptsize(–8.0)} \\
& KWT-U
  & 38.1 {\scriptsize(0.0)} 
  & 8.0 {\scriptsize(+4.3)} 
  & 71.9 {\scriptsize(–6.4)} \\
\bottomrule
\end{tabular}
}
\caption{
Comparison of alternative weighting strategies (KWT-RF and KWT-U) relative to the KWT baseline.
}
\label{tab:weighting_diff}
\vspace{-3mm}
\end{table}

\begin{table}[t]
\centering
\resizebox{0.48\textwidth}{!}{
\begin{tabular}{l l c c c}
\toprule
\textbf{Dataset} & \textbf{Model}
& \textbf{nAUPC $\uparrow$}
& \textbf{A-FPR $\downarrow$}
& \textbf{\makecell{\textbf{IDK} \\ \textbf{Precision}} $\uparrow$} \\
\midrule

% ===================== HaluEval =====================
HaluEval
& KWT + k
  & 57.8 {\scriptsize(+46.5)}
  & 3.5  {\scriptsize(–7.9)}
  & 68.6 {\scriptsize(–23.5)} \\
\midrule

% ===================== SciQ =====================
SciQ
& KWT + k
  & 62.9 {\scriptsize(+24.8)}
  & 1.4  {\scriptsize(–2.3)}
  & 81.4 {\scriptsize(+3.1)} \\
\bottomrule
\end{tabular}
}
\caption{
Effect of knowledge-based prompting on KWT.
Here, $k$ denotes external knowledge provided in the input.
}
\label{tab:k_comparison}
\end{table}

\section{Analysis of IDK expression in KWT}
\subsection{Effect of External Knowledge on IDK expression}
\label{sec:4.7}
We investigate how the model’s IDK expression changes when external knowledge is explicitly provided at inference time.
When relevant knowledge is provided in the prompt, the model should be able to answer questions that would otherwise be unanswerable based solely on its parametric knowledge.
To incorporate the external knowledge, we modify the test-time prompts following the format described in Appendix~\ref{sec:input_format}.

Table~\ref{tab:k_comparison} presents the results when knowledge is added to the input.
The results show a substantial increase in nAUPC, along with a sharp decrease in A-FPR.
IDK~Precision drops on HaluEval but improves on SciQ.
Since KWT is trained only on SFT data, it may not fully understand or follow instructions describing the provided knowledge.
This demonstrates that KWT leverages not only its internal knowledge but also externally provided information when generating responses, leading to improved accuracy and more informed decisions on whether to output <IDK>.
Nevertheless, the model exhibits large shifts in <IDK> production once the knowledge is included in the prompt.

\begin{figure}[t]
    \centering
    \includegraphics[width=0.8\linewidth]{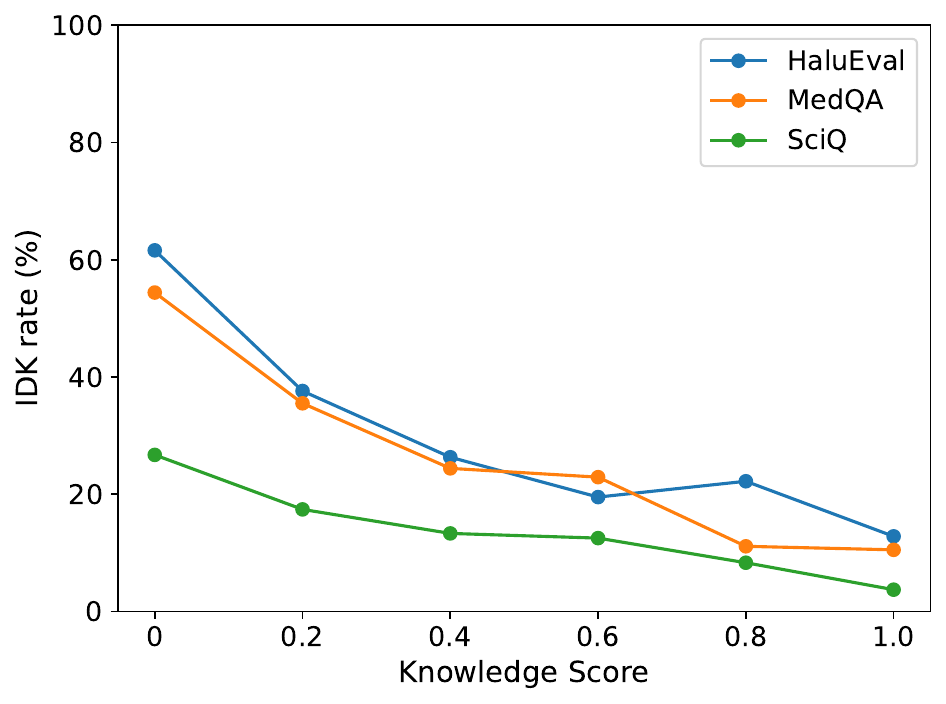}
    \caption{
    Relationship between instance-level knowledge scores and the frequency of <IDK> responses on the test dataset.
    }
    \label{fig:kwt_test_case}
    \vspace{-3mm}
\end{figure}

\subsection{Generalization of IDK expression}

Figure~\ref{fig:kwt_test_case} shows the relationship between instance-level knowledge scores and the frequency of <IDK> responses on the test dataset.
For each test instance, we estimate a knowledge score using the same procedure described in Section~\ref{sec:knowledge_estimation}, and group instances according to their scores.
For each group, we report the proportion of responses that include the <IDK> token.

As the knowledge score increases, the frequency of <IDK> responses consistently decreases.
This indicates that KWT, which is trained to express uncertainty on low-knowledge training instances, exhibits the same behavior on previously unseen test data.
The monotonic relationship indicates that the IDK expression induced during fine-tuning is consistently expressed on the test set according to the base model’s knowledge strength.

\begin{table}[t]
\centering
\resizebox{1.0\linewidth}{!}{
\small
\begin{tabular}{llccc|c}
\toprule
Backbone & Method & HaluEval & MedQA & SciQ & Avg \\
\midrule
\multirow{3}{*}{Llama3.2-3B}
& SFT & \textbf{31.0} & 20.4 & 62.7 & 38.0 \\
& KWT (non-IDK) & 30.5 & \textbf{21.1} & \textbf{65.9} & \textbf{39.2} \\
& KWT & 29.8 & 18.5 & 62.6 & 37.0 \\
\midrule
\multirow{3}{*}{Qwen2.5-3B}
& SFT & 29.5 & 19.3 & 70.7 & 39.8 \\
& KWT (non-IDK) & \textbf{30.6} & \textbf{21.3} & 71.8 & \textbf{41.2} \\
& KWT & 29.8 & 17.8 & \textbf{72.4} & 40.0 \\
\bottomrule
\end{tabular}
}
\caption{Standard accuracy of SFT, KWT, and KWT (non-IDK). 
Removing <IDK> supervision while keeping familiarity-based weighting improves average accuracy across both backbones.}
\label{tab:kwt_no_idk}
% \vspace{-3mm}
\end{table}

\begin{table}[t]
\centering
\resizebox{0.8\linewidth}{!}{
\begin{tabular}{lccc|c}
\toprule
\textbf{Model} & \textbf{HaluEval} & \textbf{MedQA} & \textbf{SciQ} & \textbf{Avg} \\
\midrule
SFT  & 3.54 & 3.06 & 5.14 & 3.91 \\
SEAL & 3.71 & 3.20 & \textbf{5.11} & 4.01 \\
KWT  & \textbf{3.30} & \textbf{2.56} & 5.20 & \textbf{3.69} \\
\bottomrule
\end{tabular}
}
\caption{Token-level KL divergence between the base model and each fine-tuned model. Lower values indicate closer alignment to the base model's predictive distribution.}
\label{tab:kl_divergence}
\vspace{-3mm}
\end{table}

\subsection{Impact of <IDK> Supervision on Accuracy Trade-off}
\label{app:idk_supervision}
To analyze the impact of <IDK> supervision on standard accuracy, we conduct an ablation study in which <IDK> is removed from the target responses while retaining the familiarity-based weighting strategy (denoted as KWT (non-IDK)). We compare this setting with standard SFT and the full KWT model, as summarized in Table~\ref{tab:kwt_no_idk}.

Across both backbones, removing <IDK> supervision consistently improves standard accuracy under the same weighting scheme. In particular, KWT (non-IDK) achieves higher average accuracy than both SFT and KWT.
This improvement suggests that familiarity-based weighting suppresses low-knowledge or high-conflict instances, allowing the model to focus on well-aligned examples and thereby improving standard accuracy compared to uniform SFT.
These results indicate that the primary source of accuracy degradation is not the instance-level weighting strategy itself, but the supervision involving the <IDK> token, which introduces a trade-off between calibrated uncertainty expression and standard accuracy. 
We suggest first training the model with SFT and then applying KWT to better balance accuracy and uncertainty, and leave this for future work.

\begin{figure}[t]
    \centering
    \begin{subfigure}[t]{0.48\linewidth}
        \centering
        \includegraphics[width=\linewidth]{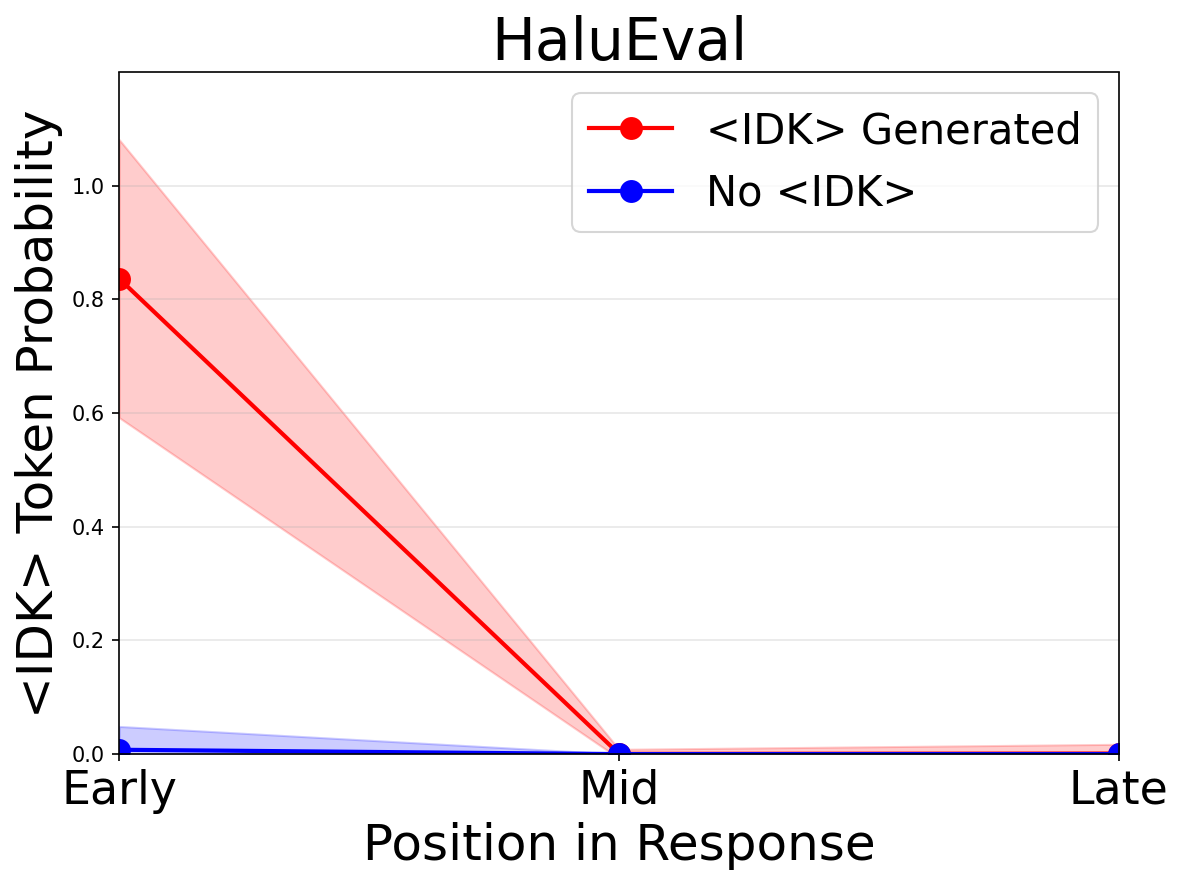}
        \caption{KWT (prepend-IDK)}
        \label{fig:kwt_idk_prepend_halueval_only}
    \end{subfigure}
    \hfill
    \begin{subfigure}[t]{0.48\linewidth}
        \centering
        \includegraphics[width=\linewidth]{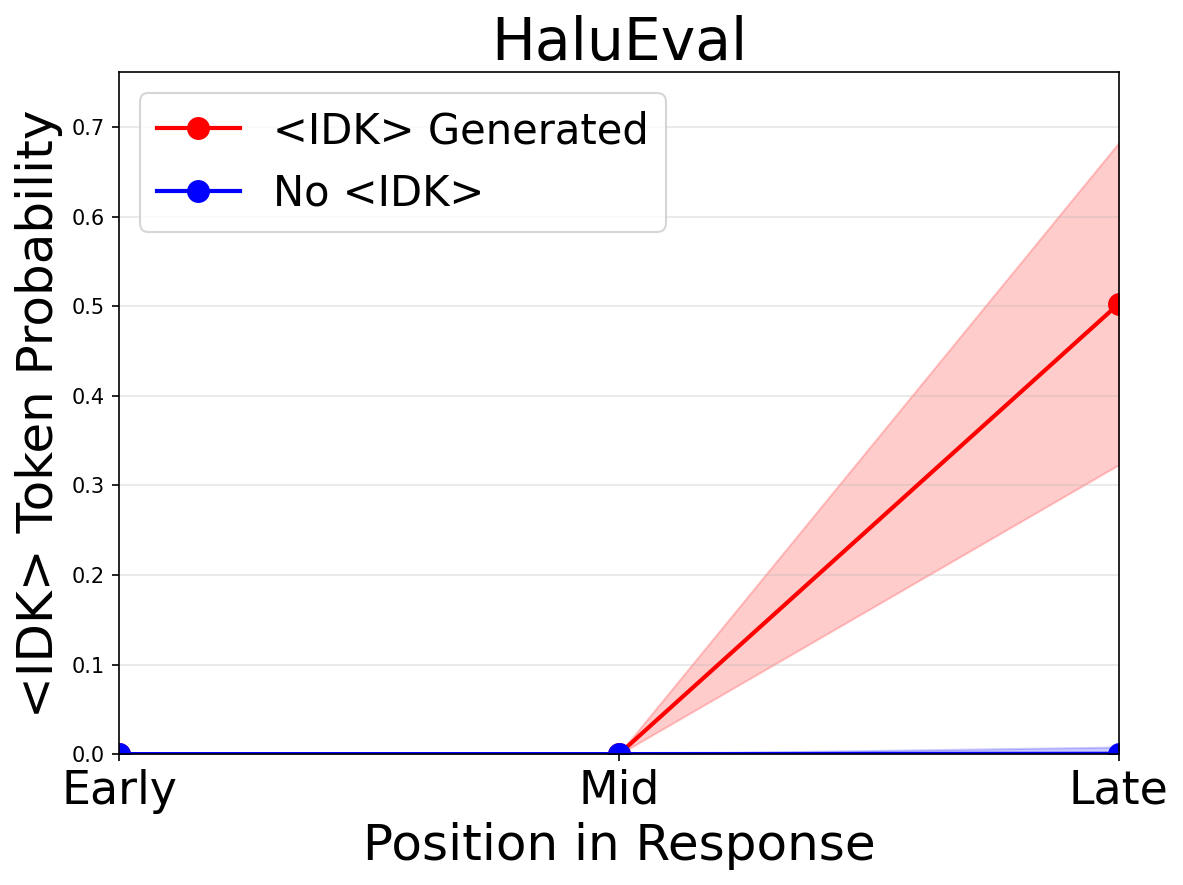}
        \caption{KWT (append-IDK)}
        \label{fig:kwt_idk_append_halueval_only}
    \end{subfigure}

    \caption{
    Token-level probability of generating the <IDK> token across relative response positions on \textbf{HaluEval}.
    The left panel shows the prepend-IDK setting, while the right panel corresponds to the append-IDK setting.
    }
    \label{fig:kwt_idk_position_halueval}
    \vspace{-1mm}
\end{figure}

\subsection{Timing of IDK Expression}
\label{sec:timing_IDK}
We adopt Append-IDK as the default strategy, where the <IDK> token is appended to the end of the response. As alternative designs, we consider Prepend-IDK, which places the token at the beginning of the response, and Only-IDK, which replaces the entire answer with the <IDK> token. Among these variants, Append-IDK consistently achieves better overall performance. A detailed comparison of these response variants is provided in Appendix~\ref{app:var_response}.

Figure~\ref{fig:kwt_idk_position_halueval} shows that the position of the <IDK> token substantially affects uncertainty expression during generation.
With Append-IDK, the probability of generating <IDK> remains near zero until the final stage, indicating that uncertainty is expressed only after the model has processed most of the response.
In contrast, Prepend-IDK concentrates <IDK> probability at the beginning of generation, resulting in an early and irreversible commitment to uncertainty expression.
As a result, Append-IDK allows the model to delay the uncertainty decision until most of the response is generated, enabling more informed uncertainty judgments.
This delayed decision is consistent with prior findings showing that later parts of a generation can reveal hallucinations, as erroneous content often persists into future context~\citep{future}.

\subsection{Predictive Distribution Shift}
We analyze how SFT, KWT, and SEAL shift the predictive distribution away from the base model by measuring the token-level KL divergence for each question–answer pair.
Specifically, for each question–answer pair, we compute the KL divergence as follows:

{\small
\setlength{\abovedisplayskip}{1pt}
\begin{equation}
\begin{aligned}
&\mathrm{KL}\!\left(M_{\text{base}} \,\|\, M_{\text{trained}}\right) \\
&= \frac{1}{T}
\sum_{t=1}^{T}
\sum_{x}
P_{\text{base}}(x \mid x_{<t}) 
\log 
\frac{
P_{\text{base}}(x \mid x_{<t})
}{
P_{\text{trained}}(x \mid x_{<t})
}.
\end{aligned}
\end{equation}
}

Table~\ref{tab:kl_divergence} reports the results averaged across all samples.
On average, KWT achieves lower KL divergence than SFT, indicating that it preserves the base model’s distribution more faithfully during fine-tuning.
This suggests that KWT avoids distorting the base model’s predictive behavior, even while improving uncertainty expression.
SEAL, in contrast, exhibits higher KL divergence than SFT, reflecting its token-level probability reallocation mechanism, which introduces larger deviations from the base distribution.

SRFT~\citep{fu2025srftsinglestagemethodsupervised} shows that large deviations from the base policy can destabilize training, highlighting the importance of maintaining proximity to the base distribution. 
Accordingly, KWT preserves the base model’s predictive distribution more effectively than both SFT and SEAL, helping maintain the model’s underlying knowledge while enabling more stable fine-tuning.

\section{Conclusion}
We show that accurately estimating how well a model knows each training instance helps mitigate knowledge misalignment between pre-training and supervised fine-tuning.
Using fine-grained knowledge scores obtained via multi-sampled inference, Knowledge-Weighted Fine-Tuning (KWT) weights the training signal to reinforce known examples while encouraging <IDK> responses on unknown ones.
This improves uncertainty expression without hurting accuracy on answerable questions, reducing hallucinations in both in-domain and out-of-domain settings.
We also introduce nAUPC, an uncertainty-aware metric that jointly evaluates accuracy and abstention behavior.
Overall, aligning fine-tuning with a model’s fine-grained knowledge offers a simple and generalizable approach to reducing hallucinations.

\section*{Limitations}
% 큰 모델에서 비용문제
Our approach relies on multi-sampled inference with multiple few-shot configurations to estimate instance-level knowledge scores, which requires performing a large number of inference runs per training instance. This inevitably increases computational cost compared to standard fine-tuning. However, this overhead is not unique to our method and is shared by many uncertainty- or consistency-based approaches that depend on repeated inference or sampling. Reducing the number of samples or developing more efficient approximations for knowledge estimation remains an important direction for future work.

\bibliography{anthology,custom}

@inproceedings{rtuning,
    title = "{R}-Tuning: Instructing Large Language Models to Say `{I} Don{'}t Know'",
    author = "Zhang, Hanning  and
      Diao, Shizhe  and
      Lin, Yong  and
      Fung, Yi  and
      Lian, Qing  and
      Wang, Xingyao  and
      Chen, Yangyi  and
      Ji, Heng  and
      Zhang, Tong",
    editor = "Duh, Kevin  and
      Gomez, Helena  and
      Bethard, Steven",
    booktitle = "Proceedings of the 2024 Conference of the North American Chapter of the Association for Computational Linguistics: Human Language Technologies (Volume 1: Long Papers)",
    month = jun,
    year = "2024",
    address = "Mexico City, Mexico",
    publisher = "Association for Computational Linguistics",
    url = "https://aclanthology.org/2024.naacl-long.394/",
    doi = "10.18653/v1/2024.naacl-long.394",
    pages = "7113--7139",
}

@inproceedings{popular,
author = {Ghosal, Gaurav and Hashimoto, Tatsunori and Raghunathan, Aditi},
title = {Understanding finetuning for factual knowledge extraction},
year = {2024},
publisher = {JMLR.org},
booktitle = {Proceedings of the 41st International Conference on Machine Learning},
articleno = {623},
numpages = {19},
location = {Vienna, Austria},
series = {ICML'24}
}

@inproceedings{seal,
    title = "Alleviating Hallucinations from Knowledge Misalignment in Large Language Models via Selective Abstention Learning",
    author = "Huang, Lei  and
      Feng, Xiaocheng  and
      Ma, Weitao  and
      Fan, Yuchun  and
      Feng, Xiachong  and
      Gu, Yuxuan  and
      Ye, Yangfan  and
      Zhao, Liang  and
      Zhong, Weihong  and
      Wang, Baoxin  and
      Wu, Dayong  and
      Hu, Guoping  and
      Kong, Lingpeng  and
      Xiao, Tong  and
      Liu, Ting  and
      Qin, Bing",
    editor = "Che, Wanxiang  and
      Nabende, Joyce  and
      Shutova, Ekaterina  and
      Pilehvar, Mohammad Taher",
    booktitle = "Proceedings of the 63rd Annual Meeting of the Association for Computational Linguistics (Volume 1: Long Papers)",
    month = jul,
    year = "2025",
    address = "Vienna, Austria",
    publisher = "Association for Computational Linguistics",
    url = "https://aclanthology.org/2025.acl-long.1199/",
    doi = "10.18653/v1/2025.acl-long.1199",
    pages = "24564--24579",
    ISBN = "979-8-89176-251-0",
}

@inproceedings{kang-etal-2025-unfamiliar,
    title = "Unfamiliar Finetuning Examples Control How Language Models Hallucinate",
    author = "Kang, Katie  and
      Wallace, Eric  and
      Tomlin, Claire  and
      Kumar, Aviral  and
      Levine, Sergey",
    editor = "Chiruzzo, Luis  and
      Ritter, Alan  and
      Wang, Lu",
    booktitle = "Proceedings of the 2025 Conference of the Nations of the Americas Chapter of the Association for Computational Linguistics: Human Language Technologies (Volume 1: Long Papers)",
    month = apr,
    year = "2025",
    address = "Albuquerque, New Mexico",
    publisher = "Association for Computational Linguistics",
    url = "https://aclanthology.org/2025.naacl-long.183/",
    doi = "10.18653/v1/2025.naacl-long.183",
    pages = "3600--3612",
    ISBN = "979-8-89176-189-6",
}

@inproceedings{NEURIPS2024_14c018d2,
 author = {Cohen, Roi and Dobler, Konstantin and Biran, Eden and de Melo, Gerard },
 booktitle = {Advances in Neural Information Processing Systems},
 doi = {10.52202/079017-0349},
 editor = {A. Globerson and L. Mackey and D. Belgrave and A. Fan and U. Paquet and J. Tomczak and C. Zhang},
 pages = {10935--10958},
 publisher = {Curran Associates, Inc.},
 title = {I Don\textquotesingle t Know: Explicit Modeling of Uncertainty with an [IDK] Token},
 url = {https://proceedings.neurips.cc/paper_files/paper/2024/file/14c018d2e72c521605b0567029ef0efb-Paper-Conference.pdf},
 volume = {37},
 year = {2024}
}

@inproceedings{gekhman-etal-2024-fine,
    title = "Does Fine-Tuning {LLM}s on New Knowledge Encourage Hallucinations?",
    author = "Gekhman, Zorik  and
      Yona, Gal  and
      Aharoni, Roee  and
      Eyal, Matan  and
      Feder, Amir  and
      Reichart, Roi  and
      Herzig, Jonathan",
    editor = "Al-Onaizan, Yaser  and
      Bansal, Mohit  and
      Chen, Yun-Nung",
    booktitle = "Proceedings of the 2024 Conference on Empirical Methods in Natural Language Processing",
    month = nov,
    year = "2024",
    address = "Miami, Florida, USA",
    publisher = "Association for Computational Linguistics",
    url = "https://aclanthology.org/2024.emnlp-main.444/",
    doi = "10.18653/v1/2024.emnlp-main.444",
    pages = "7765--7784",
}

@inproceedings{halueval,
    title = "{H}alu{E}val: A Large-Scale Hallucination Evaluation Benchmark for Large Language Models",
    author = "Li, Junyi  and
      Cheng, Xiaoxue  and
      Zhao, Xin  and
      Nie, Jian-Yun  and
      Wen, Ji-Rong",
    editor = "Bouamor, Houda  and
      Pino, Juan  and
      Bali, Kalika",
    booktitle = "Proceedings of the 2023 Conference on Empirical Methods in Natural Language Processing",
    month = dec,
    year = "2023",
    address = "Singapore",
    publisher = "Association for Computational Linguistics",
    url = "https://aclanthology.org/2023.emnlp-main.397/",
    doi = "10.18653/v1/2023.emnlp-main.397",
    pages = "6449--6464",
}

@article{medqa,
  title={What disease does this patient have? a large-scale open domain question answering dataset from medical exams},
  author={Jin, Di and Pan, Eileen and Oufattole, Nassim and Weng, Wei-Hung and Fang, Hanyi and Szolovits, Peter},
  journal={Applied Sciences},
  volume={11},
  number={14},
  pages={6421},
  year={2021},
  publisher={MDPI}
}

@inproceedings{sciq,
    title = "Crowdsourcing Multiple Choice Science Questions",
    author = "Welbl, Johannes  and
      Liu, Nelson F.  and
      Gardner, Matt",
    editor = "Derczynski, Leon  and
      Xu, Wei  and
      Ritter, Alan  and
      Baldwin, Tim",
    booktitle = "Proceedings of the 3rd Workshop on Noisy User-generated Text",
    month = sep,
    year = "2017",
    address = "Copenhagen, Denmark",
    publisher = "Association for Computational Linguistics",
    url = "https://aclanthology.org/W17-4413/",
    doi = "10.18653/v1/W17-4413",
    pages = "94--106",
}

@misc{necrefunq,
      title={Examining LLMs' Uncertainty Expression Towards Questions Outside Parametric Knowledge}, 
      author={Genglin Liu and Xingyao Wang and Lifan Yuan and Yangyi Chen and Hao Peng},
      year={2024},
      eprint={2311.09731},
      archivePrefix={arXiv},
      primaryClass={cs.CL}
}

@inproceedings{selfaware,
    title = "Do Large Language Models Know What They Don{'}t Know?",
    author = "Yin, Zhangyue  and
      Sun, Qiushi  and
      Guo, Qipeng  and
      Wu, Jiawen  and
      Qiu, Xipeng  and
      Huang, Xuanjing",
    editor = "Rogers, Anna  and
      Boyd-Graber, Jordan  and
      Okazaki, Naoaki",
    booktitle = "Findings of the Association for Computational Linguistics: ACL 2023",
    month = jul,
    year = "2023",
    address = "Toronto, Canada",
    publisher = "Association for Computational Linguistics",
    url = "https://aclanthology.org/2023.findings-acl.551/",
    doi = "10.18653/v1/2023.findings-acl.551",
    pages = "8653--8665",
}

@misc{pik,
      title={Language Models (Mostly) Know What They Know}, 
      author={Saurav Kadavath and Tom Conerly and Amanda Askell and Tom Henighan and Dawn Drain and Ethan Perez and Nicholas Schiefer and Zac Hatfield-Dodds and Nova DasSarma and Eli Tran-Johnson and Scott Johnston and Sheer El-Showk and Andy Jones and Nelson Elhage and Tristan Hume and Anna Chen and Yuntao Bai and Sam Bowman and Stanislav Fort and Deep Ganguli and Danny Hernandez and Josh Jacobson and Jackson Kernion and Shauna Kravec and Liane Lovitt and Kamal Ndousse and Catherine Olsson and Sam Ringer and Dario Amodei and Tom Brown and Jack Clark and Nicholas Joseph and Ben Mann and Sam McCandlish and Chris Olah and Jared Kaplan},
      year={2022},
      eprint={2207.05221},
      archivePrefix={arXiv},
      primaryClass={cs.CL},
      url={https://arxiv.org/abs/2207.05221}, 
}

@inproceedings{
tian2024finetuning,
title={Fine-Tuning Language Models for Factuality},
author={Katherine Tian and Eric Mitchell and Huaxiu Yao and Christopher D Manning and Chelsea Finn},
booktitle={The Twelfth International Conference on Learning Representations},
year={2024},
url={https://openreview.net/forum?id=WPZ2yPag4K}
}

@inproceedings{
wang2023selfconsistency,
title={Self-Consistency Improves Chain of Thought Reasoning in Language Models},
author={Xuezhi Wang and Jason Wei and Dale Schuurmans and Quoc V Le and Ed H. Chi and Sharan Narang and Aakanksha Chowdhery and Denny Zhou},
booktitle={The Eleventh International Conference on Learning Representations },
year={2023},
url={https://openreview.net/forum?id=1PL1NIMMrw}
}

@inproceedings{
kuhn2023semantic,
title={Semantic Uncertainty: Linguistic Invariances for Uncertainty Estimation in Natural Language Generation},
author={Lorenz Kuhn and Yarin Gal and Sebastian Farquhar},
booktitle={The Eleventh International Conference on Learning Representations },
year={2023},
url={https://openreview.net/forum?id=VD-AYtP0dve}
}

@inproceedings{
yang2024alignment,
title={Alignment for Honesty},
author={Yuqing Yang and Ethan Chern and Xipeng Qiu and Graham Neubig and Pengfei Liu},
booktitle={The Thirty-eighth Annual Conference on Neural Information Processing Systems},
year={2024},
url={https://openreview.net/forum?id=67K3Xlvw8L}
}

@misc{kalai2025languagemodelshallucinate,
      title={Why Language Models Hallucinate}, 
      author={Adam Tauman Kalai and Ofir Nachum and Santosh S. Vempala and Edwin Zhang},
      year={2025},
      eprint={2509.04664},
      archivePrefix={arXiv},
      primaryClass={cs.CL},
      url={https://arxiv.org/abs/2509.04664}, 
}

@inproceedings{li-etal-2025-know,
    title = "Know the Unknown: An Uncertainty-Sensitive Method for {LLM} Instruction Tuning",
    author = "Li, Jiaqi  and
      Tang, Yixuan  and
      Yang, Yi",
    editor = "Che, Wanxiang  and
      Nabende, Joyce  and
      Shutova, Ekaterina  and
      Pilehvar, Mohammad Taher",
    booktitle = "Findings of the Association for Computational Linguistics: ACL 2025",
    month = jul,
    year = "2025",
    address = "Vienna, Austria",
    publisher = "Association for Computational Linguistics",
    url = "https://aclanthology.org/2025.findings-acl.153/",
    doi = "10.18653/v1/2025.findings-acl.153",
    pages = "2972--2989",
    ISBN = "979-8-89176-256-5",
}

@inproceedings{li-etal-2023-contrastive,
    title = "Contrastive Decoding: Open-ended Text Generation as Optimization",
    author = "Li, Xiang Lisa  and
      Holtzman, Ari  and
      Fried, Daniel  and
      Liang, Percy  and
      Eisner, Jason  and
      Hashimoto, Tatsunori  and
      Zettlemoyer, Luke  and
      Lewis, Mike",
    editor = "Rogers, Anna  and
      Boyd-Graber, Jordan  and
      Okazaki, Naoaki",
    booktitle = "Proceedings of the 61st Annual Meeting of the Association for Computational Linguistics (Volume 1: Long Papers)",
    month = jul,
    year = "2023",
    address = "Toronto, Canada",
    publisher = "Association for Computational Linguistics",
    url = "https://aclanthology.org/2023.acl-long.687/",
    doi = "10.18653/v1/2023.acl-long.687",
    pages = "12286--12312",
}

@inproceedings{chuang2024dola,
  title={DoLa: Decoding by Contrasting Layers Improves Factuality in Large Language Models},
  author={Yung-Sung Chuang and Yujia Xie and Hongyin Luo and Yoon Kim and James R. Glass and Pengcheng He},
  booktitle={The Twelfth International Conference on Learning Representations},
  year={2024},
  url={https://openreview.net/forum?id=Th6NyL07na}
}

@inproceedings{shi-etal-2024-trusting,
    title = "Trusting Your Evidence: Hallucinate Less with Context-aware Decoding",
    author = "Shi, Weijia  and
      Han, Xiaochuang  and
      Lewis, Mike  and
      Tsvetkov, Yulia  and
      Zettlemoyer, Luke  and
      Yih, Wen-tau",
    editor = "Duh, Kevin  and
      Gomez, Helena  and
      Bethard, Steven",
    booktitle = "Proceedings of the 2024 Conference of the North American Chapter of the Association for Computational Linguistics: Human Language Technologies (Volume 2: Short Papers)",
    month = jun,
    year = "2024",
    address = "Mexico City, Mexico",
    publisher = "Association for Computational Linguistics",
    url = "https://aclanthology.org/2024.naacl-short.69/",
    doi = "10.18653/v1/2024.naacl-short.69",
    pages = "783--791",
}

@inproceedings{
madaan2023selfrefine,
title={Self-Refine: Iterative Refinement with Self-Feedback},
author={Aman Madaan and Niket Tandon and Prakhar Gupta and Skyler Hallinan and Luyu Gao and Sarah Wiegreffe and Uri Alon and Nouha Dziri and Shrimai Prabhumoye and Yiming Yang and Shashank Gupta and Bodhisattwa Prasad Majumder and Katherine Hermann and Sean Welleck and Amir Yazdanbakhsh and Peter Clark},
booktitle={Thirty-seventh Conference on Neural Information Processing Systems},
year={2023},
url={https://openreview.net/forum?id=S37hOerQLB}
}

@inproceedings{
mndler2024selfcontradictory,
title={Self-contradictory Hallucinations of Large Language Models: Evaluation, Detection and Mitigation},
author={Niels M{\"u}ndler and Jingxuan He and Slobodan Jenko and Martin Vechev},
booktitle={The Twelfth International Conference on Learning Representations},
year={2024},
url={https://openreview.net/forum?id=EmQSOi1X2f}
}

@misc{chen2023universalselfconsistency,
      title={Universal Self-Consistency for Large Language Model Generation}, 
      author={Xinyun Chen and Renat Aksitov and Uri Alon and Jie Ren and Kefan Xiao and Pengcheng Yin and Sushant Prakash and Charles Sutton and Xuezhi Wang and Denny Zhou},
      year={2023},
      eprint={2311.17311},
      archivePrefix={arXiv},
      primaryClass={cs.CL},
      url={https://arxiv.org/abs/2311.17311}, 
}

@inproceedings{wang-etal-2024-integrate,
    title = "Integrate the Essence and Eliminate the Dross: Fine-Grained Self-Consistency for Free-Form Language Generation",
    author = "Wang, Xinglin  and
      Li, Yiwei  and
      Feng, Shaoxiong  and
      Yuan, Peiwen  and
      Pan, Boyuan  and
      Wang, Heda  and
      Hu, Yao  and
      Li, Kan",
    editor = "Ku, Lun-Wei  and
      Martins, Andre  and
      Srikumar, Vivek",
    booktitle = "Proceedings of the 62nd Annual Meeting of the Association for Computational Linguistics (Volume 1: Long Papers)",
    month = aug,
    year = "2024",
    address = "Bangkok, Thailand",
    publisher = "Association for Computational Linguistics",
    url = "https://aclanthology.org/2024.acl-long.634/",
    doi = "10.18653/v1/2024.acl-long.634",
    pages = "11782--11794",
}

@inproceedings{wang-etal-2024-improving,
    title = "Improving {LLM} Generations via Fine-Grained Self-Endorsement",
    author = "Wang, Ante  and
      Song, Linfeng  and
      Peng, Baolin  and
      Jin, Lifeng  and
      Tian, Ye  and
      Mi, Haitao  and
      Su, Jinsong  and
      Yu, Dong",
    editor = "Ku, Lun-Wei  and
      Martins, Andre  and
      Srikumar, Vivek",
    booktitle = "Findings of the Association for Computational Linguistics: ACL 2024",
    month = aug,
    year = "2024",
    address = "Bangkok, Thailand",
    publisher = "Association for Computational Linguistics",
    url = "https://aclanthology.org/2024.findings-acl.499/",
    doi = "10.18653/v1/2024.findings-acl.499",
    pages = "8424--8436",
}

@misc{future,
      title={Enhancing Hallucination Detection via Future Context}, 
      author={Joosung Lee and Cheonbok Park and Hwiyeol Jo and Jeonghoon Kim and Joonsuk Park and Kang Min Yoo},
      year={2025},
      eprint={2507.20546},
      archivePrefix={arXiv},
      primaryClass={cs.CL},
      url={https://arxiv.org/abs/2507.20546}, 
}

@misc{grattafiori2024llama3herdmodels,
      title={The Llama 3 Herd of Models}, 
      author={Aaron Grattafiori and others},
      year={2024},
      eprint={2407.21783},
      archivePrefix={arXiv},
      primaryClass={cs.AI},
      url={https://arxiv.org/abs/2407.21783}, 
}

@misc{gemmateam2025gemma3technicalreport,
      title={Gemma 3 Technical Report}, 
      author={{Gemma Team and Google DeepMind}},
      year={2025},
      eprint={2503.19786},
      archivePrefix={arXiv},
      primaryClass={cs.CL},
      url={https://arxiv.org/abs/2503.19786}, 
}

@misc{qwen2025qwen25technicalreport,
      title={Qwen2.5 Technical Report}, 
      author={{Qwen Team}},
      year={2025},
      eprint={2412.15115},
      archivePrefix={arXiv},
      primaryClass={cs.CL},
      url={https://arxiv.org/abs/2412.15115}, 
}

@inproceedings{zhong-etal-2025-kaft,
    title = "{K}a{FT}: Knowledge-aware Fine-tuning for Boosting {LLM}s' Domain-specific Question-Answering Performance",
    author = "Zhong, Qihuang  and
      Ding, Liang  and
      Cai, Xiantao  and
      Liu, Juhua  and
      Du, Bo  and
      Tao, Dacheng",
    editor = "Che, Wanxiang  and
      Nabende, Joyce  and
      Shutova, Ekaterina  and
      Pilehvar, Mohammad Taher",
    booktitle = "Findings of the Association for Computational Linguistics: ACL 2025",
    month = jul,
    year = "2025",
    address = "Vienna, Austria",
    publisher = "Association for Computational Linguistics",
    url = "https://aclanthology.org/2025.findings-acl.1235/",
    doi = "10.18653/v1/2025.findings-acl.1235",
    pages = "24085--24100",
    ISBN = "979-8-89176-256-5",
}

@inproceedings{ren-etal-2024-learning,
    title = "Learning or Self-aligning? Rethinking Instruction Fine-tuning",
    author = "Ren, Mengjie  and
      Cao, Boxi  and
      Lin, Hongyu  and
      Liu, Cao  and
      Han, Xianpei  and
      Zeng, Ke  and
      Guanglu, Wan  and
      Cai, Xunliang  and
      Sun, Le",
    editor = "Ku, Lun-Wei  and
      Martins, Andre  and
      Srikumar, Vivek",
    booktitle = "Proceedings of the 62nd Annual Meeting of the Association for Computational Linguistics (Volume 1: Long Papers)",
    month = aug,
    year = "2024",
    address = "Bangkok, Thailand",
    publisher = "Association for Computational Linguistics",
    url = "https://aclanthology.org/2024.acl-long.330/",
    doi = "10.18653/v1/2024.acl-long.330",
    pages = "6090--6105",
}

@misc{fu2025srftsinglestagemethodsupervised,
      title={SRFT: A Single-Stage Method with Supervised and Reinforcement Fine-Tuning for Reasoning}, 
      author={Yuqian Fu and Tinghong Chen and Jiajun Chai and Xihuai Wang and Songjun Tu and Guojun Yin and Wei Lin and Qichao Zhang and Yuanheng Zhu and Dongbin Zhao},
      year={2025},
      eprint={2506.19767},
      archivePrefix={arXiv},
      primaryClass={cs.CL},
      url={https://arxiv.org/abs/2506.19767}, 
}

@article{zheng2023judging,
  title={Judging llm-as-a-judge with mt-bench and chatbot arena},
  author={Zheng, Lianmin and Chiang, Wei-Lin and Sheng, Ying and Zhuang, Siyuan and Wu, Zhanghao and Zhuang, Yonghao and Lin, Zi and Li, Zhuohan and Li, Dacheng and Xing, Eric and others},
  journal={Advances in neural information processing systems},
  volume={36},
  pages={46595--46623},
  year={2023}
}
\bibliographystyle{acl_natbib}

\appendix

\section{Prompt Templates}
\subsection{Input Format}
\label{sec:input_format}
During training, the input format is as follows:
\begin{tcolorbox}[colback=gray!10, colframe=black!75, title=Training format]
\small
Question: \texttt{\{question\}}

Answer: \texttt{\{response\}}
\end{tcolorbox}

Here, the response is set to the gold answer if the knowledge score is greater than zero, and to the gold answer followed by <IDK> if the knowledge score is zero.
As described in Section~\ref{sec:4.7}, when external knowledge is provided at inference time, the prompt is modified by adding a Knowledge: {knowledge} field before the question.

\subsection{LLM-as-a-Judge Evaluation Prompt}
\label{sec:llm-as-a-judeg-prompt}
To compute the knowledge score, we must determine whether each model-generated response matches the gold answer.
One approach is to use an LLM-as-a-judge framework, which evaluates correctness using the following prompt:
\begin{tcolorbox}[colback=gray!10, colframe=black!75, title=Matching Prompt]
\small
Question: \texttt{\{question\}}

Knowledge: \texttt{\{knowledge\}}

Answer1: \texttt{\{gold answer\}}

Answer2: \texttt{\{model answer\}}

Are Answer1 and Answer2 semantically equivalent?

Answer only 'yes' or 'no':
\end{tcolorbox}
Here, the \texttt{Knowledge} field is included only when the dataset provides an associated supporting knowledge paragraph; otherwise, it is omitted from the prompt.
This method is also used for evaluating the model performance.

\section{IDK Ratio}
\label{app:idk_ratio}

\paragraph{IDK Score.}
The OOD evaluation datasets provide explicit labels indicating whether each question is answerable or unanswerable.
Since unanswerable questions should ideally trigger <IDK> response, while answerable questions should not, this metric combines the IDK Rate (IR) on both subsets.
It assigns higher scores to models that produce fewer <IDK> on answerable queries and more <IDK> on unanswerable ones.

{\small
\begin{equation}
\mathrm{IR} = \frac{100\times(A + B)}{A + B + C + D}
\end{equation}
}

{\small
\begin{equation}
\mathrm{IDK\ Score} = \frac{(100 - \mathrm{IR}_{\text{ans}}) + \mathrm{IR}_{\text{unans}}}{2}
\end{equation}
}

Table~\ref{tab:ood_idk} shows the proportion of <IDK> responses. 
KWT achieves the highest IDK score on unanswerable questions in both NEC and RefuNQ, indicating more appropriate uncertainty-aware behavior. 
In contrast, R-Tuning consistently produces very high <IDK> ratios, even on answerable questions, showing a tendency to over-refuse rather than distinguish between answerable and unanswerable inputs.
This pattern remains the same in out-of-domain evaluations, indicating that precise knowledge scoring is crucial for improving both accuracy and uncertainty-aware behavior across settings.

\section{Experiments with Qwen 2.5}
\label{sec:qwen2.5}
Table~\ref{tab:qwen2.5} reports the results using Qwen~2.5-3B as the base model ($M_{\text{base}}$), and the overall trends are consistent with our main experiments.
Across all datasets, KWT (LLM) achieves the highest nAUPC while substantially reducing A-FPR and improving IDK Precision compared to all baselines.
Although KWT (LLM) attains slightly lower standard accuracy than R-Tuning, it remains comparable to SFT and other KWT variants, indicating that improved uncertainty expression is achieved without severe accuracy degradation.
In contrast, SEAL exhibits perfect A-FPR and IDK Precision due to deterministic token-level abstention, but suffers from substantial accuracy degradation across datasets.

\begin{table*}[t]
\centering
\resizebox{1.0\textwidth}{!}{
    \begin{tabular}{lccccccccc}
    \toprule
    \textbf{Model} 
    & \multicolumn{3}{c}{\textbf{NEC}} 
    & \multicolumn{3}{c}{\textbf{RefuNQ}} 
    & \multicolumn{3}{c}{\textbf{SelfAware}} \\
    \cmidrule(lr){2-4} \cmidrule(lr){5-7} \cmidrule(lr){8-10}
    & \textbf{$\boldsymbol{\mathrm{IR}_{\text{ans}} \downarrow}$} & \textbf{$\boldsymbol{\mathrm{IR}_{\text{unans}} \uparrow}$} & \textbf{IDK Score $\uparrow$}
    & \textbf{$\boldsymbol{\mathrm{IR}_{\text{ans}} \downarrow}$} & \textbf{$\boldsymbol{\mathrm{IR}_{\text{unans}} \uparrow}$} & \textbf{IDK Score $\uparrow$}
    & \textbf{$\boldsymbol{\mathrm{IR}_{\text{ans}} \downarrow}$} & \textbf{$\boldsymbol{\mathrm{IR}_{\text{unans}} \uparrow}$} & \textbf{IDK Score $\uparrow$} \\
    \midrule
    
    R-Tuning 
    & 94.45 & 100    & 52.78 
    & 87.82 & 95.44  & 53.81 
    & 86.95 & 77.71  & \textbf{45.38} \\
    
    % SEAL 
    % & 28.88 & 46.49  & 58.81 
    % & 26.74 & 47.03  & 60.15 
    % & 27.17 & 20.54  & 46.69 \\
    
    KWT (EM) 
    & 79.92 & 81.67  & 50.88 
    & 68.62 & 81.96  & 56.67 
    & 68.46 & 43.02  & 37.28 \\
    
    KWT (Rouge)
    & 38.18 & 49.76  & 55.79 
    & 50.57 & 69.86  & 59.65 
    & 39.54 & 21.22  & 40.84 \\
    
    \rowcolor{lightgray}
    KWT (LLM)
    & 50.05 & 67.13  & \textbf{58.54}
    & 47.13 & 66.77  & \textbf{59.82} 
    & 41.98 & 16.18  & 37.10 \\
    \bottomrule
    \end{tabular}
}
\caption{Comparison of IDK ratios on the out-of-domain datasets: NEC, RefuNQ, and SelfAware.}
\label{tab:ood_idk}
\end{table*}

\begin{table*}[t]
\centering
\resizebox{0.7\textwidth}{!}{
\begin{tabular}{l l c c c c}
\toprule
\textbf{Dataset} & \textbf{Model} 
& \textbf{Acc $\uparrow$} 
& \textbf{nAUPC $\uparrow$} 
& \textbf{A-FPR $\downarrow$} 
& \textbf{IDK Precision $\uparrow$} \\
\midrule

\multirow{7}{*}{HaluEval}
& R-Tuning    & 30.4 & 5.7 & 38.3 & 83.5 \\
& SEAL        & 27.8 & 10.6 & 0.0$^\dagger$ & 100.0$^\dagger$ \\
& KWT (EM)    & \textbf{30.9} & 9.1 & 25.6 & 87.0 \\
& KWT (Rouge) & \textbf{30.9} & 10.9 & 14.9 & 89.5 \\
& KWT-RF (LLM) & 30.1 & 11.0 & 17.0 & 90.7 \\
& KWT-U (LLM) & 30.1 & 10.9 & 15.6 & 90.5 \\
\rowcolor{lightgray}
& KWT (LLM)   & 29.8 & \textbf{11.3} & \textbf{6.5} & \textbf{93.6} \\
\midrule

\multirow{7}{*}{MedQA}
& R-Tuning    & \textbf{20.0} & -2.4 & 78.0 & 82.5 \\
& SEAL        & 17.7 & \textbf{5.7} & 0.0$^\dagger$ & 100.0$^\dagger$ \\
& KWT (EM)    & 18.7 & -0.6 & 66.0 & 83.9 \\
& KWT (Rouge) & 16.9 & 1.6 & 46.5 & 88.3 \\
& KWT-RF (LLM) & 20.6 & 1.9 & 47.7 & 86.4 \\
& KWT-U (LLM) & 19.2 & 2.4 & 41.8 & 87.8 \\
\rowcolor{lightgray}
& KWT (LLM)   & 17.8 & 3.9 & \textbf{24.7} & \textbf{91.2} \\
\midrule

\multirow{7}{*}{SciQ}
& R-Tuning    & \textbf{73.5} & 22.2 & 32.4 & 45.8 \\
& SEAL        & 70.7 & 51.2 & 0.0$^\dagger$ & 100.0$^\dagger$ \\
& KWT (EM)    & 70.6 & 32.5 & 19.4 & 55.1 \\
& KWT (Rouge) & 70.8 & 41.4 & 10.7 & 63.6 \\
& KWT-RF (LLM) & 69.4 & 42.3 & 8.7 & 69.1 \\
& KWT-U (LLM) & 70.8 & 46.3 & 6.4 & 72.6 \\
\rowcolor{lightgray}
& KWT (LLM)   & 72.4 & \textbf{51.3} & \textbf{2.8} & \textbf{78.3} \\
\bottomrule
\end{tabular}
}
\caption{Comparison of uncertainty-aware behavior on the in-domain datasets using the Qwen~2.5 model.
$^\dagger$ indicates a deterministic outcome induced by token-level abstention, where any output containing <IDK> is counted as incorrect.}
\label{tab:qwen2.5}
\end{table*}

\section{Variation of Response}
\label{app:var_response}
We adopt the append strategy as our default approach, where an <IDK> token is added at the end of the response when the knowledge score is zero.
Table~\ref{tab:kwt_idk_variants} reports the results for alternative strategies, where <IDK> is either prepended to the response or used to fully replace the original answer.

KWT (append-IDK) outperforms KWT (prepend-IDK) across all metrics. This suggests that allowing the model to first produce an answer and then express uncertainty yields more stable behavior.
KWT (only-IDK), which replaces the entire response with <IDK>, behaves similarly to SEAL in that any output containing <IDK> is counted as incorrect. As a result, it achieves perfect A-FPR and IDK Precision but suffers from substantial accuracy degradation.

\section{Additional Analysis on Hallucination}

To directly analyze hallucination behavior, we measure how often the model generates <IDK> among incorrect responses using the following recall metric:

{\small
\begin{equation}
\mathrm{IDK\text{-}Recall} = \frac{B}{B + D}
\end{equation}
}
where $B$ denotes incorrect answers that include <IDK>, and $D$ denotes incorrect answers without <IDK>. 
A higher value indicates that the model more frequently expresses uncertainty when it produces an incorrect response.

\begin{table}[t]
\centering
\small
\begin{tabular}{lccc}
\toprule
Model & HaluEval & MedQA & SciQ \\
\midrule
R-Tuning & \textbf{89.5} & \textbf{89.6} & \textbf{73.7} \\
SEAL & 35.0 & 33.8 & 12.4 \\
KWT (LLM) & 56.1 & 51.1 & 22.2 \\
\bottomrule
\end{tabular}
\caption{Hallucination recall ($B/(B+D)$) across datasets.}
\label{tab:hallucination_recall}
\vspace{-3mm}
\end{table}

R-Tuning achieves the highest recall; however, when considered with A-FPR and IDK-Precision, this largely reflects excessive <IDK> generation rather than well-calibrated uncertainty. This shows that recall alone does not necessarily indicate effective hallucination mitigation.
Since metrics such as FPR or recall in isolation are insufficient for selecting the optimal model, we argue that nAUPC—capturing the trade-off between uncertainty-aware abstention and certainty preservation—provides a more comprehensive evaluation criterion.

\begin{table*}[t]
\centering
\resizebox{0.7\linewidth}{!}{
\begin{tabular}{l l c c c c}
\toprule
\textbf{Dataset} & \textbf{Model} 
& \textbf{Acc $\uparrow$} 
& \textbf{nAUPC $\uparrow$} 
& \textbf{A-FPR $\downarrow$} 
& \textbf{IDK Precision $\uparrow$} \\
\midrule

\multirow{3}{*}{HaluEval}
& KWT (prepend-IDK) & 29.7 & 8.3 & 26.8 & 87.2 \\
& KWT (only-IDK)    & 21.2 & 11.2 & 0.0$^\dagger$ & 100.0$^\dagger$ \\
& KWT (append-IDK)  & \textbf{29.8} & \textbf{11.3} & \textbf{11.4} & \textbf{92.1} \\
\midrule

\multirow{3}{*}{MedQA}
& KWT (prepend-IDK) & 16.3 & 1.7 & 45.2 & 89.6 \\
& KWT (only-IDK)    & 7.5  & 3.6 & 0.0$^\dagger$ & 100.0$^\dagger$ \\
& KWT (append-IDK)  & \textbf{18.5} & \textbf{4.3} & \textbf{21.6} & \textbf{91.2} \\
\midrule

\multirow{3}{*}{SciQ}
& KWT (prepend-IDK) & \textbf{62.8} & 29.8 & 15.6 & 65.6 \\
& KWT (only-IDK)    & 52.7 & 35.3 & 0.0$^\dagger$ & 100.0$^\dagger$ \\
& KWT (append-IDK)  & 62.6 & \textbf{38.1} & \textbf{3.7} & \textbf{78.3} \\
\bottomrule
\end{tabular}}
\caption{Comparison of KWT variants across datasets with both standard accuracy and nAUPC.
KWT (prepend-IDK) prepends the <IDK> token to the response, KWT (only-IDK) replaces the entire response with <IDK>, and KWT (append-IDK) appends <IDK> to the end of the response, which is the default setting in our experiments.
$^\dagger$ indicates a deterministic outcome induced by token-level abstention, where any output containing <IDK> is counted as incorrect.}
\label{tab:kwt_idk_variants}
\end{table*}

\begin{figure*}[t]
    \centering

    % ===================== Prepend <IDK> =====================
    \begin{subfigure}[t]{0.32\linewidth}
        \centering
        \includegraphics[width=\linewidth]{figure/kwt_ridk_halueval_idk_prob_sequence_no_interp.png}
        \caption{KWT (prepend-IDK) in HaluEval}
        \label{fig:kwt_idk_prepend_halueval}    
    \end{subfigure}
    \hfill
    \begin{subfigure}[t]{0.32\linewidth}
        \centering
        \includegraphics[width=\linewidth]{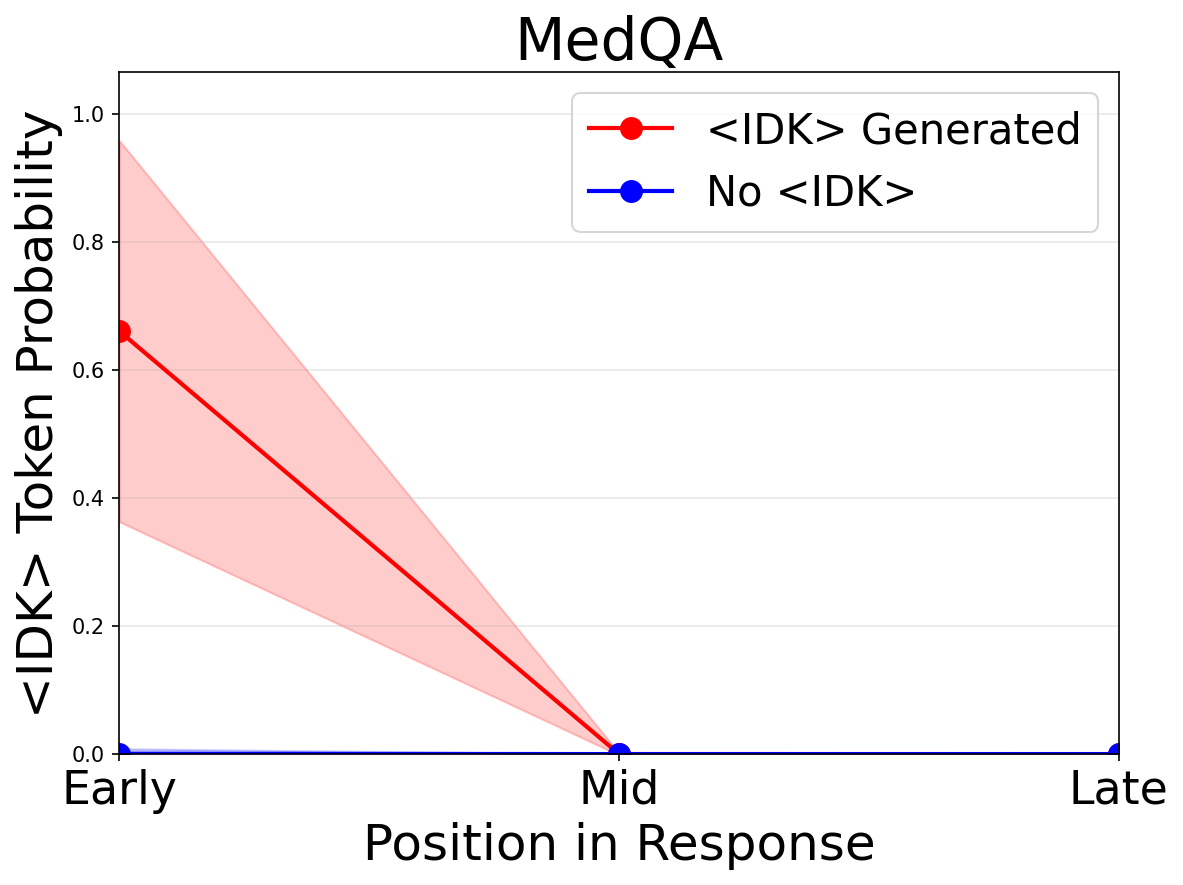}
        \caption{KWT (prepend-IDK) in MedQA}
        \label{fig:kwt_idk_prepend_medqa}
    \end{subfigure}
    \hfill
    \begin{subfigure}[t]{0.32\linewidth}
        \centering
        \includegraphics[width=\linewidth]{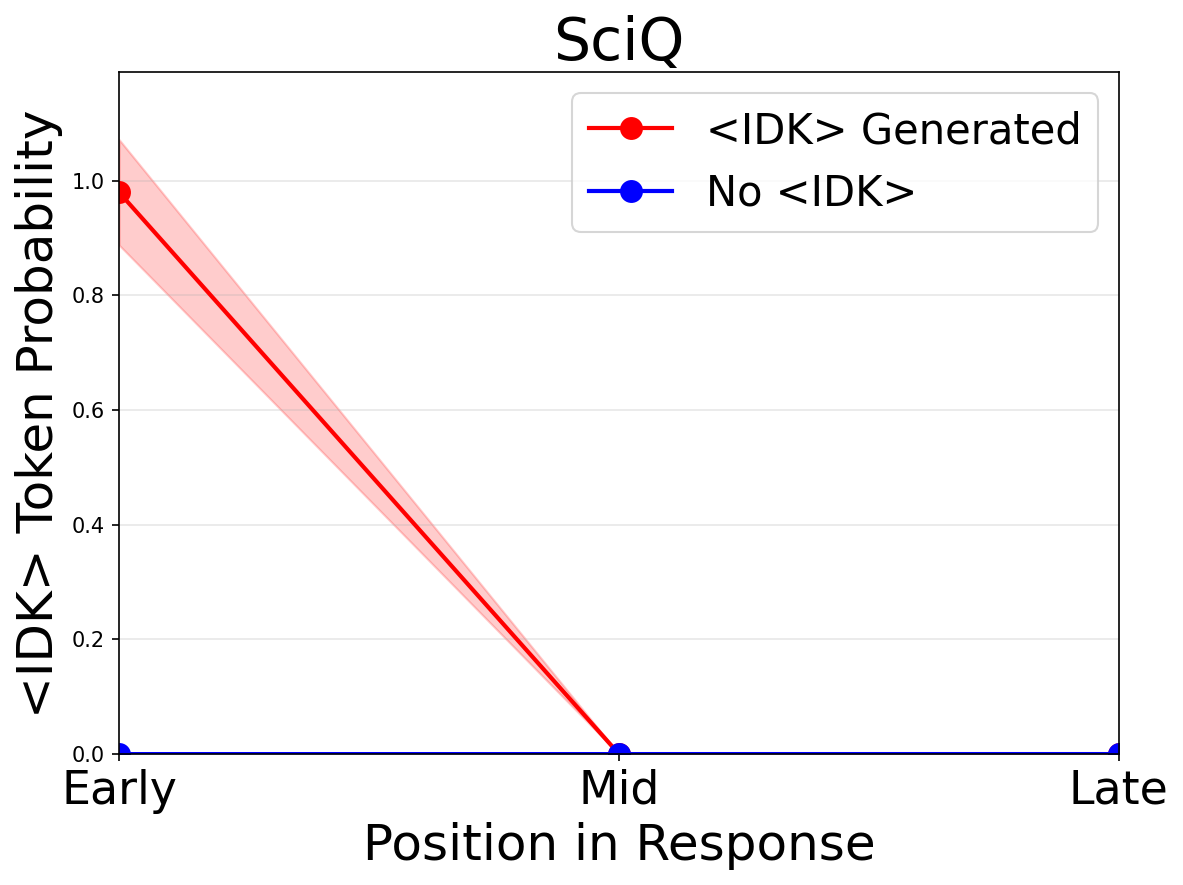}
        \caption{KWT (prepend-IDK) in SciQ}
        \label{fig:kwt_idk_prepend_sciq}
    \end{subfigure}

    \vspace{2mm}

    % ===================== Append <IDK> =====================
    \begin{subfigure}[t]{0.32\linewidth}
        \centering
        \includegraphics[width=\linewidth]{figure/kwt_halueval_idk_prob_sequence_no_interp.png}
        \caption{KWT (append-IDK) in HaluEval}
        \label{fig:kwt_idk_append_halueval}
    \end{subfigure}
    \hfill
    \begin{subfigure}[t]{0.32\linewidth}
        \centering
        \includegraphics[width=\linewidth]{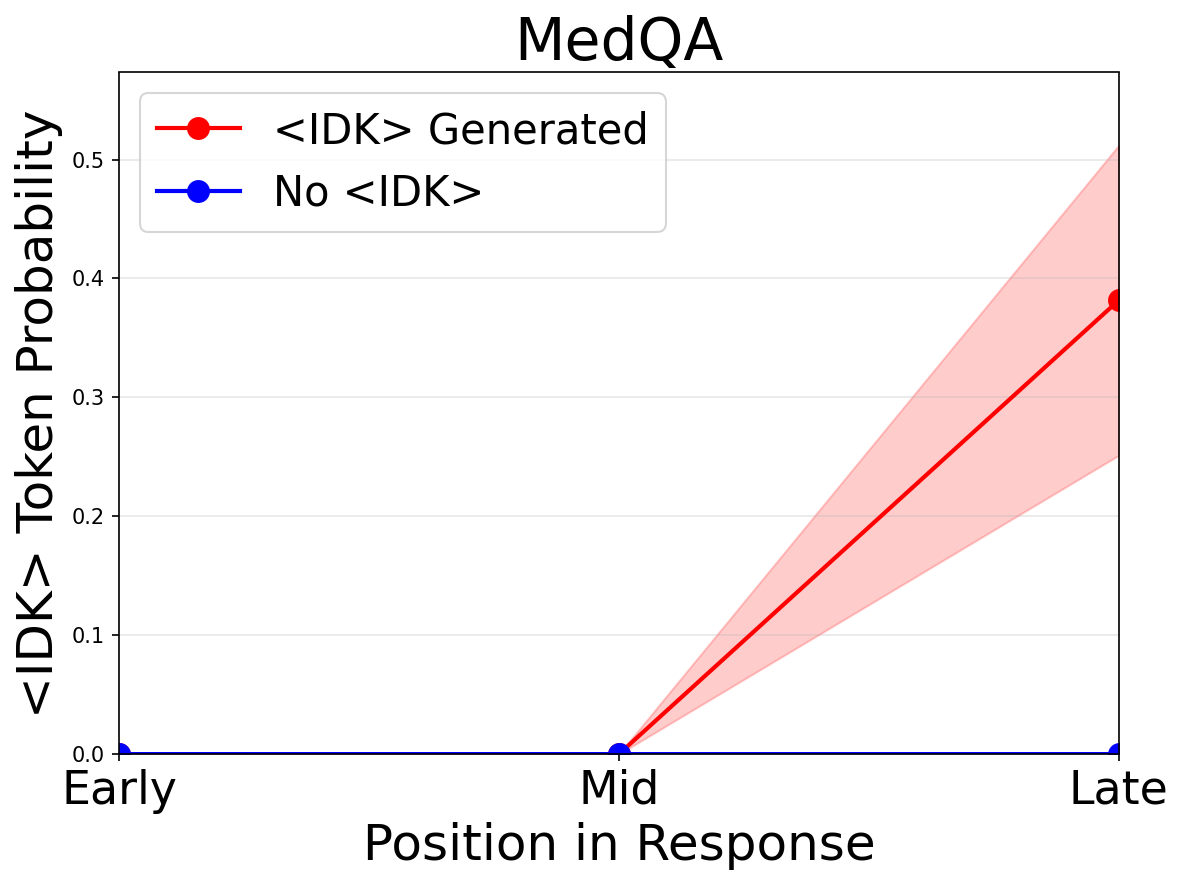}
        \caption{KWT (append-IDK) in MedQA}
        \label{fig:kwt_idk_append_medqa}
    \end{subfigure}
    \hfill
    \begin{subfigure}[t]{0.32\linewidth}
        \centering
        \includegraphics[width=\linewidth]{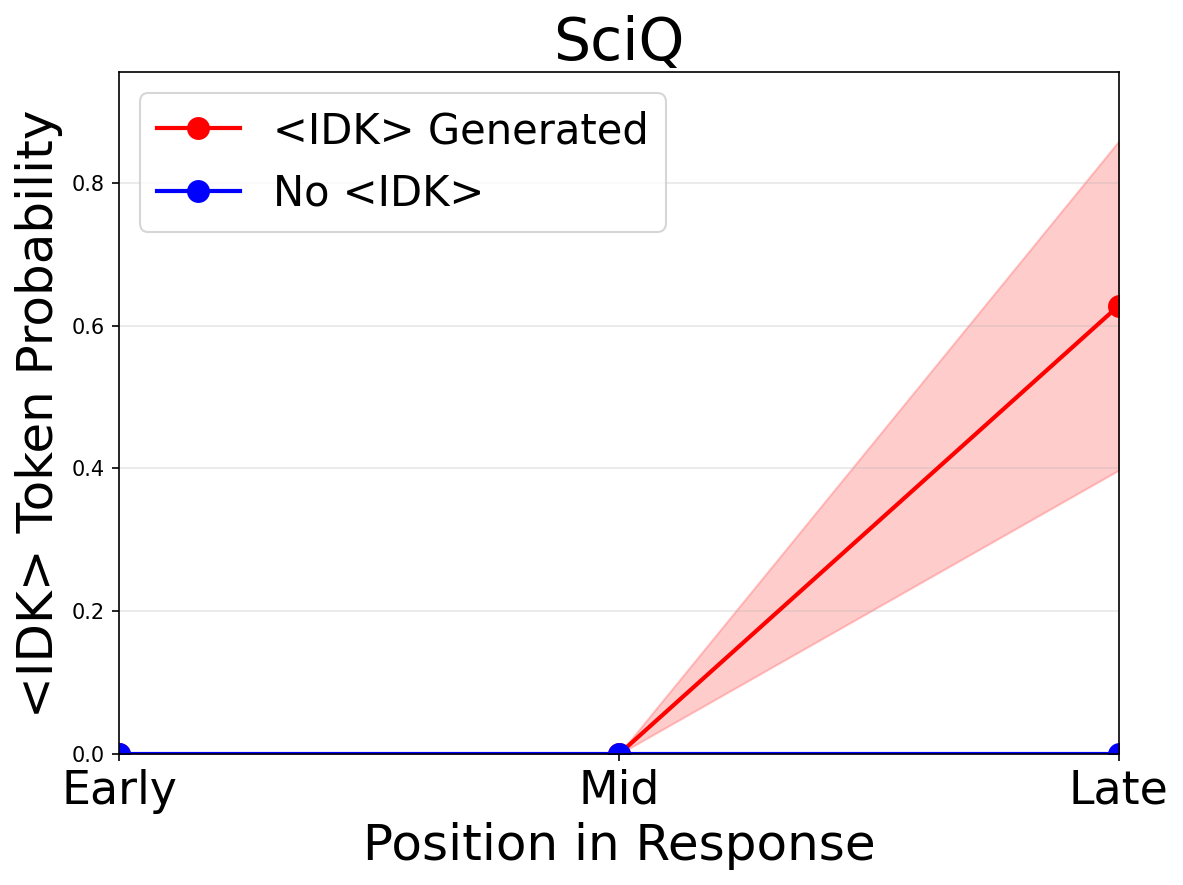}
        \caption{KWT (append-IDK) in SciQ}
        \label{fig:kwt_idk_append_sciq}
    \end{subfigure}    
    \caption{
    Token-level probability of generating the <IDK> token at each relative position in the response under KWT.
    The top row corresponds to the prepend-IDK setting, while the bottom row shows the append-IDK setting.
    }
    \label{fig:kwt_idk_position}
    \vspace{-3mm}
\label{fig:kwt_idk_position}
\end{figure*}

\end{document}